\documentclass[10pt,twocolumn]{article}

\usepackage[a4paper,margin=0.85in]{geometry}
\usepackage{times}
\usepackage{amsmath}
\usepackage{amssymb}
\usepackage{microtype}
\usepackage[utf8]{inputenc}
\usepackage[T1]{fontenc}
\usepackage{graphicx}
\usepackage{xcolor}
\usepackage{booktabs}
\usepackage{multirow}
\usepackage{listings}
\usepackage{url}
\usepackage[hidelinks]{hyperref}
\usepackage{enumitem}
\usepackage{authblk}
\usepackage{abstract}
\usepackage{titlesec}
\usepackage{caption}
\usepackage{subcaption}
\usepackage{stfloats}

\titleformat{\section}{\large\bfseries}{\thesection.}{0.5em}{}
\titleformat{\subsection}{\normalsize\bfseries}{\thesubsection}{0.5em}{}

\setlist{nosep, leftmargin=*}

\newcommand{\specsec}[1]{UB-spec~\S#1}

\title{\textbf{OpenURMA: A Clean-Room Open Implementation of \\
the Unified Bus Protocol}}

\author{Bojie Li \\ Pine AI}

\date{}

\begin{document}

\twocolumn[
  \begin{@twocolumnfalse}
    \maketitle
    \vspace{-2em}
    \begin{abstract}
      \noindent
      Modern datacenter RDMA is bottlenecked at the network interface,
      not the wire. A NIC running RoCE or InfiniBand holds
      per-connection state for every (application, remote-endpoint)
      pair --- hundreds of megabytes at 1024-application fanout ---
      and pays a four-traversal PCIe round trip on a 64-byte
      operation, inflating latency an order of magnitude beyond the
      wire. Both follow from the Queue-Pair-over-PCIe abstraction
      RDMA inherits from InfiniBand.

      Huawei's \emph{Unified Bus} (UB), a public 2025 specification,
      changes the abstraction: it decouples per-application endpoint
      state from per-host transport state so connection context grows
      additively, exposes ordering as opt-in, and reaches remote
      memory through native CPU load/store to an on-chip-bus
      controller. UB ships in Huawei's closed Ascend~950 silicon.

      \emph{OpenURMA} is the first clean-room open implementation of
      UB's transport and transaction layers, realised at three tiers
      --- synthesisable RTL on Alveo~U50, a cycle-level two-node
      SystemC simulator, and a gem5 full-system scaffold --- each with
      a matched OpenRoCE (RoCEv2~RC) baseline. The contribution is the
      implementation, harness, and controlled comparison closed silicon
      does not admit. On the canonical 64-byte remote fetch --- LOAD
      on \specsec{8.3}, READ on RoCEv2~RC --- UB's load/store path delivers
      $\approx$500~ns end-to-end, \textbf{4.37$\times$} below the
      matched baseline (2186~ns), sustains 2.80$\times$ higher
      throughput, and fits in $\approx$14\% of a U50's LUTs.
    \end{abstract}
    \vspace{1em}
  \end{@twocolumnfalse}
]

\section{Introduction}
\label{sec:intro}

A modern datacenter network is two interconnects fused into one.
From the application's view, RDMA looks like a load or store to
remote memory --- bus-like, low-latency, address-and-go. From the
NIC's view, every operation is a queued work request shuttled
across PCIe to a peripheral device --- network-like, with all of
the network's per-message machinery and the peripheral's PCIe
crossing. For two decades the fusion was tractable: connection
counts stayed small, operation sizes stayed large, and the gap
between the bus-like view and the network-like implementation
could be papered over with cleverer software above the device.
AI-training workloads broke both assumptions. A single job now
exchanges 64-byte gradient updates across thousands of GPUs; the
bus-like view's promises collide with the network-like
implementation's costs, and the collision is structural.

The two costs of the collision are both consequences of one design
stance: \emph{the NIC is a peripheral}. Peripherals communicate
with the CPU through a paired-queue programming model, so the NIC
must hold per-pair connection state; peripherals sit behind PCIe,
so every operation crosses the PCIe boundary. Under RoCEv2~RC,
each (local application, remote endpoint) pair is a Queue Pair
(QP) carrying $\sim$512~B of NIC state; at 1024 of each, the NIC's
working set is $\sim$537~MB, past any commodity NIC's on-chip SRAM
and into the regime where every operation pays a PCIe round trip
to refetch its
QP~\cite{kong2022:collie,kalia:fasst-extended,wang2023:srnic}.
The same operation pays a second cost even when state stays on
chip: a 64-byte READ spends roughly 1.5~$\mu$s of its $\sim$2~$\mu$s
round trip on four PCIe traversals (doorbell MMIO, work-queue
DMA fetch, completion DMA write, CPU poll-miss across PCIe
coherence), while the wire delivers in $\sim$200~ns. Software
cannot move the peripheral; hardware inside the peripheral cannot
move PCIe. What removes both costs is moving the peripheral.

Huawei's \emph{Unified Bus} (UB)~\cite{ubspec} is a 2025
specification that moves the peripheral. The NIC is no longer
behind PCIe but on the on-chip bus, addressed by the CPU through
ordinary load and store instructions; connection state is no
longer per-application-pair but per-host-pair; ordering is no
longer always-on but opt-in. Huawei's Ascend~950
NPU (neural processing unit)~\cite{huawei2026:ascend950} ships UB
at commodity scale, but
the silicon is closed and the spec is unaccompanied by a public
implementation that researchers can measure, instrument, or
prototype against.

\begin{figure*}[t]
  \centering
  \includegraphics[width=0.95\textwidth]{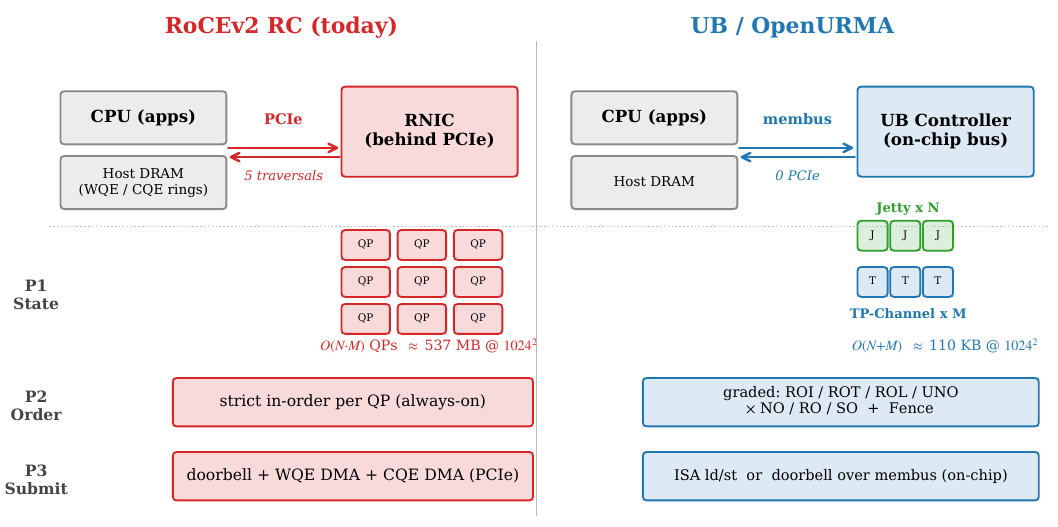}
  \caption{The three architectural moves and their dependencies.
  RoCEv2~RC (top) puts the NIC behind PCIe, holds one Queue~Pair
  per (application, remote-endpoint) pair, and enforces strict
  order on every operation. UB (bottom) splits state along the
  transaction/transport line so it grows as $O(N{+}M)$; places
  the controller on the on-chip bus, where the CPU can reach it
  with loads and stores; and exposes ordering as four opt-in axes
  that reuse counters the layer split already provisions. The
  arrows mark the chain: bounded state is what lets the controller
  live on-bus; the on-bus controller is what lets the load/store
  path exist; the layer split is what makes ordering cheap.}
  \label{fig:three-pillars}
\end{figure*}

The three architectural moves UB makes are not independent
choices; they form a chain in which each move makes the next
possible (Figure~\ref{fig:three-pillars}).

\begin{enumerate}\itemsep 3pt

\item \textbf{Split the transport layer.} The Queue Pair fused
application identity with transport reliability into one object,
so state had to scale with the product of the two counts. UB
separates the two: per-application endpoint state lives on a
\emph{Jetty}; per-remote-host transport state lives on a
\emph{TP~Channel}. Per-NIC state then grows additively in the
number of local endpoints~$N$ (one per thread at the
high-performance design point, since threads sharing a work queue
serialize on it) and remote hosts~$M$,\footnote{$M$ counts remote
hosts --- the conservative one-endpoint-per-host case. RC binds
each QP to a specific remote endpoint, so a host exposing $P$
addressable endpoints costs RoCE $N{\cdot}M{\cdot}P$ QPs; UB's
TP~Channel is per-host regardless of $P$, because the remote
application is named in the packet header rather than bound into
the connection.} as $O(N{+}M)$ instead of $O(N{\cdot}M)$. Bounding
state is what makes (2) possible.

\item \textbf{Move the controller onto the on-chip bus.} A NIC
whose working set fits in on-chip SRAM can live next to the CPU
on the on-chip bus rather than behind PCIe; a NIC whose working
set spills to host DRAM cannot, because each spill becomes a
host-memory access. Putting the controller on-bus is what makes
(3) possible.

\item \textbf{Admit a load/store data path.} Once the controller
is on the on-chip bus, a CPU's load or store instruction can
reach it directly. The four PCIe traversals collapse into a
single on-chip-bus crossing, and the work-queue and
completion-queue machinery elides for small synchronous
operations.

\end{enumerate}

\noindent A fourth move --- opt-in ordering --- rides on the same
per-application counters the first move provisions, so it costs
zero extra pipeline cycles on operations that do not request
gating, while letting workloads that need a barrier pay only for
that barrier.

OpenURMA is a clean-room open implementation of UB's transport and
transaction layers, realised at three modeling tiers --- RTL on
Alveo~U50, a two-node SystemC simulator that models both endpoints
of the wire, and a gem5 full-system scaffold running real ARM
binaries against the SystemC NIC --- each against a matched
\emph{OpenRoCE} (RoCEv2~RC) baseline on the same toolchain and
harness, so both stacks' numbers come from identical modeling
assumptions rather than published vendor data. On the canonical
64-byte remote fetch, the load/store path delivers $\approx$500~ns
end-to-end --- \textbf{4.37$\times$} below the RoCEv2 baseline
(2186~ns) for the same CPU-fetches-64~B operation.

\paragraph{Contributions.}
\begin{itemize}\itemsep 2pt
  \item \textbf{First clean-room open implementation of the
    Unified Bus protocol.} Synthesisable RTL for the transport
    and transaction layers, closing 322~MHz post-route at roughly
    14\% of an Alveo~U50's LUT budget.
  \item \textbf{Apples-to-apples OpenRoCE baseline.} A RoCEv2~RC
    implementation on the same toolchain, target, and harness.
  \item \textbf{Multi-tier evaluation infrastructure.} A
    cycle-level two-node SystemC simulator that accounts for both
    endpoints of the wire, plus a gem5 full-system scaffold that
    puts a CPU in the loop; jointly necessary to measure the
    end-to-end CPU-to-remote-DRAM path UB claims to shorten.
  \item \textbf{Cross-cutting validation,} reproducing published
    ConnectX-7 RDMA WRITE latency within $\pm$5\%, congestion-control
    dynamics against DCQCN, selective-vs-Go-Back-N loss recovery, and
    a YCSB-A application port.
\end{itemize}

\noindent OpenURMA is released at
\url{https://github.com/bojieli/OpenURMA}.

\paragraph{Roadmap.}
\S\ref{sec:bg} places UB in context against the
peripheral-NIC abstraction and three decades of patches around it.
\S\ref{sec:design} walks the design choices and \S\ref{sec:impl}
the pipeline decomposition and timing closure.
\S\ref{sec:methodology} sets up the evaluation;
\S\ref{sec:feasibility} establishes that the NIC synthesises, is
cheap, characterises its raw latency, and validates the model
against ConnectX-7. The evaluation is then
organised by design commitment. The four moves of
\S\ref{subsec:ub-changes} reduce to three measurable commitments:
bounded state (Move~1), the load/store latency collapse (Moves~2
and~3 jointly --- the on-bus controller's payoff \emph{is} the
latency at which a load/store reaches remote memory, so the two are
measured as one number), and opt-in ordering (Move~4).
\S\ref{sec:eval-state} validates bounded state,
\S\ref{sec:eval-latency} the load/store latency collapse, and
\S\ref{sec:eval-ordering} near-free opt-in ordering;
\S\ref{sec:eval-fullsystem} confirms all three under a real OS.
\S\ref{sec:eval-transport} then covers reliable transport under loss
and congestion, and \S\ref{sec:eval-scaleout} the scale-out reach
coherent fabrics cannot match, with \S\ref{sec:eval-summary}
summarising. \S\ref{sec:discussion} discusses limitations;
\S\ref{sec:related} surveys related work.

\section{Background}
\label{sec:bg}

This section grounds the introduction's thesis in concrete
mechanics: what makes the NIC a peripheral, why the peripheral
stance imposes both costs by construction, why three decades of
fixes around it leave the abstraction in place, and what Unified
Bus replaces it with.

\subsection{The peripheral-NIC abstraction}
\label{subsec:qp-cost}

Every commodity RDMA NIC --- InfiniBand HCAs, Mellanox/NVIDIA
ConnectX, Broadcom Thor, Intel IPU --- is a PCIe peripheral. The
CPU sees it through a memory-mapped BAR; the NIC sees the CPU
through DMA. The pair communicates by a Queue-Pair protocol: the
CPU posts work-queue entries to host memory, signals via MMIO
doorbells, and the NIC consumes them by DMA and writes completions
back the same way. This programming model is itself a small
network protocol spoken across PCIe, and the model fixes two
properties of the system before the wire is even reached.

\paragraph{The QP fuses what should be two layers.} A Queue~Pair
is named by a (local-application, remote-endpoint) pair and
carries both transaction-layer state (work queues, permissions,
the identity of \emph{who is calling}) and transport-layer state
(sequence numbers, retransmit window, RTO timer, congestion state
--- the bookkeeping of \emph{how packets are delivered}). Textbook
protocol design separates these layers; the QP collapses them into
one object whose unit of accounting is the application pair, not
the application count or the peer count. Total NIC state therefore
grows as $O(N{\cdot}M)$. At $(N,M){=}(1024,1024)$, $\sim$1~M QPs
at $\sim$512~B each is $\sim$537~MB, beyond any commodity NIC's
on-chip SRAM. When the working set spills to host DRAM, every
operation pays a PCIe round trip to refetch its QP before it can
even consult its sequence
number~\cite{kong2022:collie,wang2023:srnic}.

\paragraph{The PCIe attachment imposes four traversals per
operation.} Because the NIC is a peripheral, every operation
crosses the PCIe boundary four times. On a 64-byte READ: the CPU
writes a work-queue entry to host DRAM and then a doorbell to the
NIC's MMIO BAR (traversal~1); the NIC DMA-reads the work-queue
entry (traversal~2); the NIC emits the wire request, receives the
response, and DMA-writes a completion entry to host DRAM
(traversal~3); the CPU polls the completion queue and incurs a
PCIe-coherence miss to fetch the freshly written line
(traversal~4). Of the $\sim$2~$\mu$s round-trip time,
$\sim$1.5~$\mu$s is consumed by these four traversals; the wire
itself delivers in $\sim$200~ns. The traversals are not an
implementation accident; they are what the model requires when the
two communicating parties live in disjoint address spaces.

\paragraph{Both costs follow from one stance.}
Fixing either in isolation leaves the other in place: connection
sharing (XRC, the eXtended Reliable Connection, and SRQ, the
Shared Receive Queue) reduces state but keeps the four PCIe
traversals; doorbell coalescing and inline work-queue-entry (WQE)
shortcuts trim one traversal but keep the QP fusion. The abstraction is what limits
the optimisation budget.

\subsection{Why prior fixes are patches}
\label{subsec:patches}

Three decades of work on RDMA scale fall in three buckets, none
of which removes both costs.
\emph{Software above the device} ---
FaSST~\cite{kalia:fasst-extended}, 1RMA~\cite{singhvi2020:1rma}, Snap's
Pony Express~\cite{marty2019:snap}, the broader SmartNIC-pacing
literature --- works above
the verb interface; it can change the application's view but not
the peripheral attachment or the programming model.
\emph{Hardware inside the device} ---
SRNIC~\cite{wang2023:srnic}, StaR~\cite{wang2021:star}, the IRN
loss-recovery line~\cite{mittal2018:irn,lu2017:melo,huang2024:fasr,li2025:dcp},
MP-RDMA~\cite{lu2018:mprdma}, ConWeave~\cite{tan2023:conweave}
--- rebuilds the NIC's internals symptom by symptom, but the NIC
remains a PCIe peripheral with a Queue-Pair programming model.
\emph{Programmable substrates} ---
Tonic~\cite{arashloo2020:tonic},
NanoTransport~\cite{ibanez2021:nanotransport},
StRoM~\cite{strom-eurosys20} --- host new transports on FPGA or
P4 fabric, but they are substrates, not new transports.

\subsection{Unified Bus: the spec, the silicon, the gap}
\label{subsec:ub-position}

UB is the first RDMA-class specification since RoCEv2 (2014) to
replace the abstraction rather than patch it. The 2025 spec spans
physical through verb layers; this paper engages only transport
and transaction, where the structural choices live. The protocol
ships in Huawei's Ascend~950 NPU~\cite{huawei2026:ascend950}
with both the work-queue and load/store data paths as on-die
blocks alongside the AI accelerators. The user-space verb library
and kernel driver are open-source; the protocol implementation is
closed. OpenURMA fills that gap with a clean-room implementation
built from the public specification.

\subsection{Making the NIC a peer of the CPU}
\label{subsec:ub-changes}

\begin{figure*}[t]
  \centering
  \includegraphics[width=0.95\textwidth]{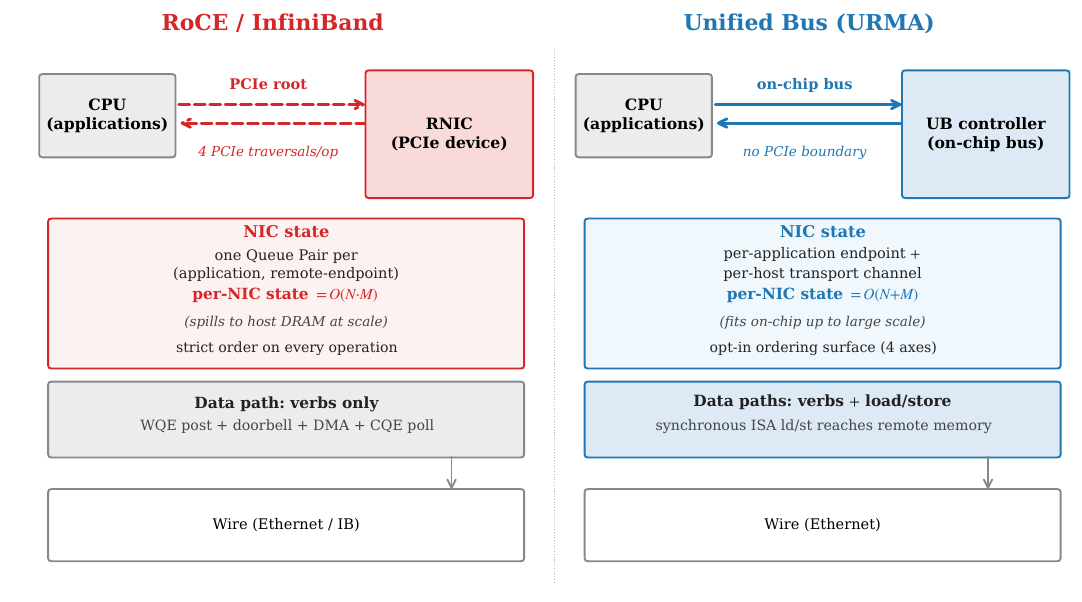}
  \caption{Architectural comparison. RoCE puts the NIC behind PCIe;
  it holds one Queue~Pair per (application, remote-endpoint) pair,
  enforces strict order on every operation, and admits only the
  verb-driven data path. Unified Bus puts the controller on the
  on-chip bus; it holds per-application endpoint state separately
  from per-remote-host transport state, exposes ordering as an
  opt-in surface, and admits CPU load/store as a synchronous data
  path alongside the verb path.}
  \label{fig:arch-comparison}
\end{figure*}

UB stops treating the NIC as a peripheral the CPU programs and
starts treating it as a peer the CPU shares an on-chip bus with
(Figure~\ref{fig:arch-comparison}). This section gives the
spec-level mechanism behind each move \S\ref{sec:intro} introduced
--- the same enabling chain (bounded state $\rightarrow$ on-bus
controller $\rightarrow$ load/store path, with opt-in ordering
riding on the first), now with the field-level detail.

\begin{figure*}[t]
  \centering
  \includegraphics[width=0.95\textwidth]{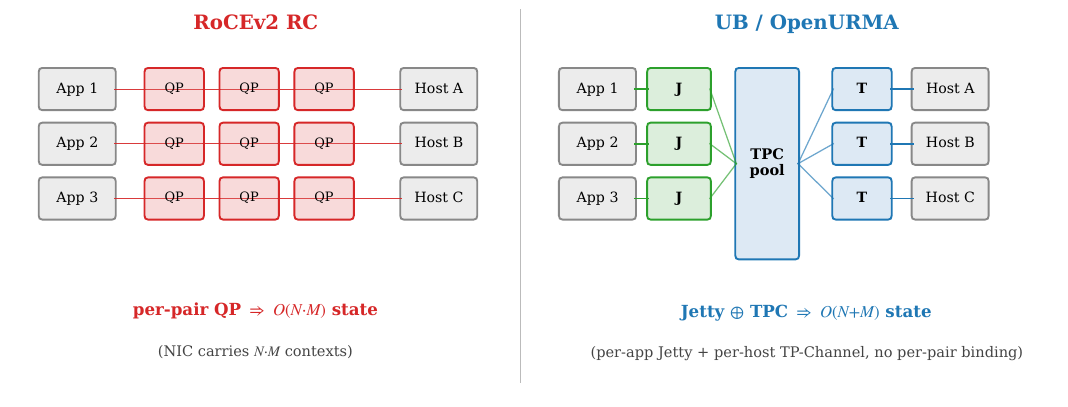}
  \caption{State models compared. RoCE binds one Queue~Pair to every
  (application, remote-host) pair, so per-NIC state grows as
  $O(N{\cdot}M)$. UB decouples per-application state (Jetty) from
  per-host transport state (TP~Channel); any Jetty addresses any
  remote endpoint through the shared TP~Channel pool, and per-NIC
  state grows as $O(N{+}M)$.}
  \label{fig:state-model}
\end{figure*}

\paragraph{Move 1: split transaction from transport.}
The QP fused two layers into one object, so state had to scale
with their product. UB separates them: a \emph{Jetty}\footnote{The name marks the
break from the connection. A \emph{jetty} is a wharf: one
structure at which many vessels berth in turn, sharing the
harbour rather than each dredging a private channel to shore.
The metaphor is deliberate --- a Jetty is an application's berth
at the network, not a wire to one peer, and the requests it
issues occupy that berth only while each is in flight. A new
abstraction wants a word that does not smuggle in the old one's
assumptions: ``connection'' and ``queue pair'' both presume a
bound peer, which is exactly what the layer split removes.}
holds the
transaction-layer state for one local application
(completion-queue handle, access token, and type/state flags,
$\sim$20~B); a \emph{TP~Channel}
holds the transport-layer state for one remote host (sequence-number
windows, SACK bitmap, congestion state, $\sim$56~B)
(Figure~\ref{fig:state-model}). Any Jetty addresses any remote
endpoint through the shared per-host TP~Channel pool, with the
destination carried in the packet header rather than baked into
a per-pair binding. State grows as $O(N{+}M)$: at
$(N,M){=}(1024,1024)$, the Jetty, TP~Channel, and shared
memory-region tables together hold $\sim$110~KB instead of 537~MB
(\S\ref{sec:eval-state}).

The split pays for itself with one extra lookup per outbound
packet (which TP~Channel by destination host) and a protocol
surface that exposes Jetty and TP~Channel as separate first-class
objects with separate lifecycles. The QP's one-object-per-peer
simplicity is the cost of admission. The split bounds the
cross-product, not per-endpoint contention: a Jetty carries its own
work queue and doorbell, so sharing one across threads serializes
exactly as a shared QP does. $N$ therefore counts independent issue
contexts identically for both stacks; what UB removes is the
$M$-fold multiplication of that count, not the per-endpoint
serialization.

\paragraph{Move 2: put the controller on the on-chip bus.}
At $O(N{+}M)$ state the connection working set fits in on-chip SRAM
at scales that previously forced spill to host DRAM, so the
controller can live on the CPU's on-chip bus rather than behind
PCIe: the CPU reaches it through ordinary memory-mapped regions and
the PCIe boundary collapses into a single bus crossing. The cost is
that the controller must sit on the CPU's bus segment.

\begin{figure*}[t]
  \centering
  \includegraphics[width=0.95\textwidth]{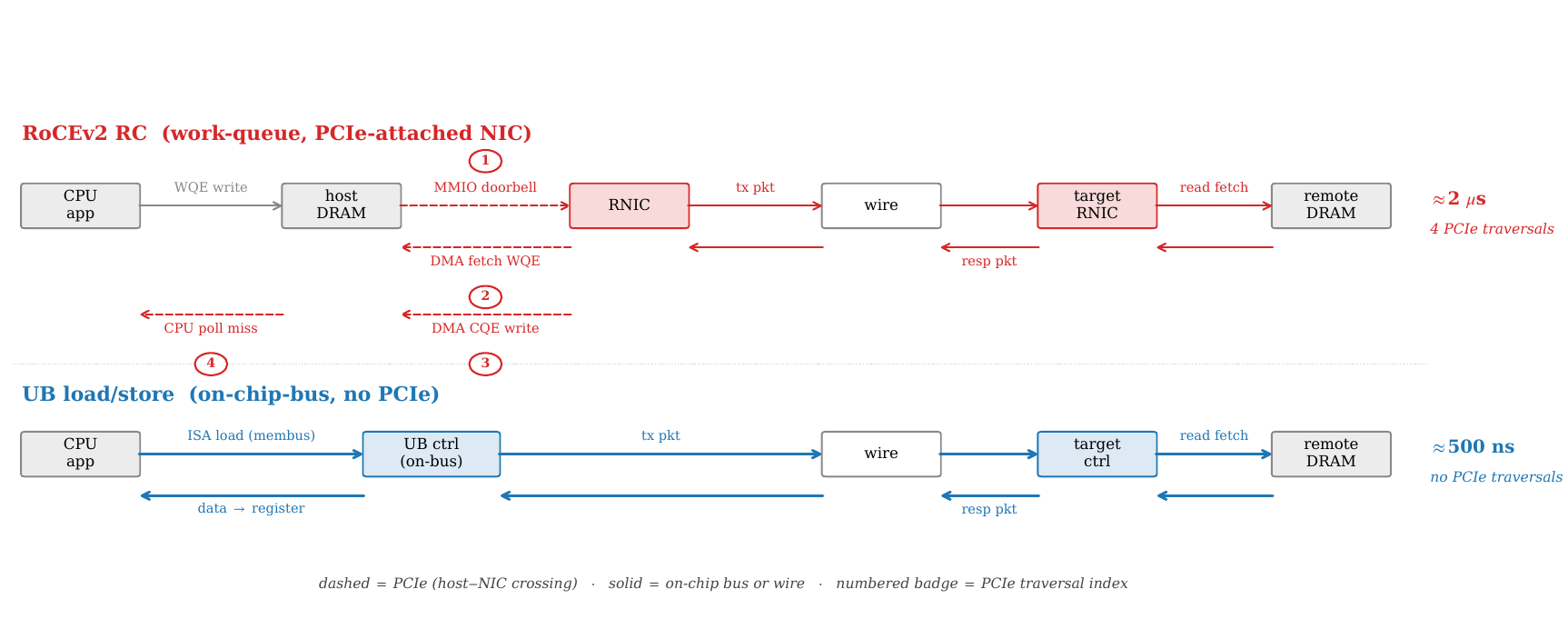}
  \caption{Per-operation data path for a small synchronous read.
  The traditional work-queue-driven path (top) traverses four PCIe
  crossings --- doorbell over MMIO, work-queue-entry DMA fetch,
  completion-entry DMA write, and a CPU poll-miss across PCIe
  coherence. The Unified Bus load/store path (bottom) replaces all
  four with on-chip-bus crossings: a CPU load instruction reaches
  the controller, the controller emits the wire packet, the
  response arrives, and the data lands directly in the CPU register
  that issued the load.}
  \label{fig:dataflow-comparison}
\end{figure*}

\paragraph{Move 3: admit a load/store data path.}
With the controller on-bus, a CPU load or store reaches an address
aperture it owns (Figure~\ref{fig:dataflow-comparison}); the
controller turns each instruction into one wire transaction --- no
work-queue entry, doorbell, DMA, or separate completion --- and the
load blocks until the response returns. Only short synchronous
operations (loads, stores, same-path atomics) are admissible; bulk
and asynchronous traffic stays on the work-queue path. The two
complement rather than compete --- load/store wins on short,
latency-tight operations, the work-queue path on long,
throughput-bound ones.

\paragraph{Move~4: make ordering opt-in.}
This move sits outside the chain but on top of Move~1. RoCE~RC
enforces strict in-order delivery on every operation; UB exposes
ordering as four orthogonal axes the application opts into per
channel and per operation. The \emph{service mode} controls how
aggressively packets may reorder on the wire (relaxed admits
multi-path spreading; strict forces single-path delivery); the
\emph{execution order} controls whether the issue stage may emit
before the previous operation commits; the \emph{fence} is an
explicit application barrier; the \emph{completion order} bit
controls whether completions retire in arrival or issue order.
A pure-fanout broadcast bypasses every gate; a
barrier-after-collective pays exactly the gating its workload
needs.

Opt-in ordering is cheap, robust, and semantically apt at once.
\emph{Cheap:} the gating indexes per-Jetty counters Move~1 already
provisions, so it adds zero pipeline cycles on operations that
bypass it. \emph{Robust:} strict total order is fragile under
failure --- a NIC promised in-order delivery cannot retire any
completion while an earlier operation stalls, so one slow path
head-of-line-blocks every application on the channel. \emph{Apt:}
collective and gradient-exchange traffic is order-insensitive by
construction (the reduction is commutative; the application
resynchronises at the barrier), so strict per-operation order
spends reliability machinery on a guarantee the workload never
asked for. The opt-in surface lets each workload pay only the
gating it needs.

\subsection{The OpenClickNP toolchain}
\label{subsec:openclicknp}

OpenURMA is built on OpenClickNP~\cite{openclicknp}, a clean-room
re-implementation of the ClickNP element model~\cite{li2016:clicknp}.
The model is a familiar one: a NIC pipeline is a directed graph of
\emph{elements}, each with typed input and output ports, local
state, and a per-cycle handler. The toolchain lowers a single
source description through three back-ends: a software emulator
for unit testing, a cycle-accurate SystemC simulator for performance
measurement, and a Vitis~HLS back-end that emits synthesisable C++
for Vivado synthesis on the Alveo~U50. The same source produces all
three artifacts. The relevance to this paper is twofold: the model
makes element-level cost (LUTs, pipeline-II, BRAM) directly
measurable, and the three back-ends are how OpenURMA produces both
the RTL tier and the SystemC tier from one description.

\section{Design}
\label{sec:design}

This section describes how OpenURMA realises Unified Bus transaction- and transport-layer protocol in RTL. We
organise the description around the three architectural commitments
the four moves of \S\ref{subsec:ub-changes} reduce to bounded
state (Move~1), opt-in ordering (Move~4), and the load/store data
path (the on-bus controller of Move~2 together with the load/store
path of Move~3).

\begin{figure*}[t]
  \centering
  \includegraphics[width=0.99\textwidth]{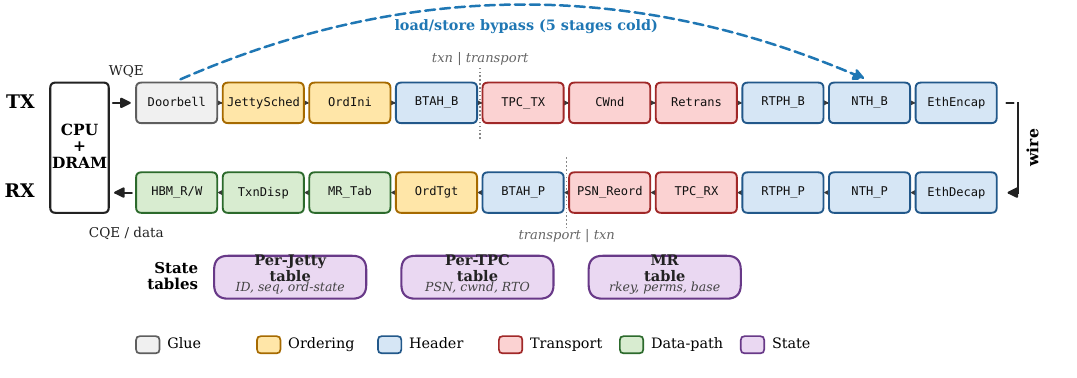}
  \caption{OpenURMA's NIC as a ClickNP element graph. The TX path
  (top) flows from CPU doorbell to wire; the RX path (bottom)
  inverts. Each box is one element, coloured by its functional
  category (legend at bottom): glue, ordering, header parse/build,
  transport, on-NIC data path, and state. The dotted vertical
  markers separate the transaction layer (application-level Jetty
  scheduling, ordering, memory-region checks, atomic operations,
  completion generation) from the transport layer (per-host
  sequencing, retransmission, congestion control). The dashed arc
  shows the load/store data path
  (\S\ref{subsec:design-loadstore}): a CPU load or store reaches
  the controller and the bypass element produces a wire packet in
  five stages cold, skipping the ordering and transport-state
  machinery entirely.}
  \label{fig:topology}
\end{figure*}

\subsection{Realising bounded state}
\label{subsec:design-state}

The Jetty / TP~Channel split maps onto two pipeline regions. The
transmit path consults a per-Jetty table to admit and gate the
work request, then hands it to a per-host TP~Channel scheduler
that attaches the sequence number, picks the multi-path lane, and
queues the packet. The receive path inverts the flow: per-TP-Channel
logic validates and reorders by sequence number, then a per-Jetty
table maps the destination header to the right receive queue.
Two tables, two indices, two scaling regimes. The pipeline pays
one extra header field per packet (the destination Jetty id) and
one extra lookup per side of the wire; in exchange, nothing in
the pipeline binds a transmit context to a receive context, so
the protocol's $O(N{+}M)$ scaling becomes the pipeline's. At
$(N,M){=}(1024,1024)$ the per-NIC connection-state working set is $\sim$110~KB
versus $\sim$537~MB for the matched RoCE QP table
(\S\ref{sec:eval-state}).

\subsection{Realising opt-in ordering}
\label{subsec:design-ordering}

The four ordering axes become four gating stages on the pipeline.
Each decides at one well-defined point whether the operation in
flight may proceed; together they form a mesh that adds zero
pipeline cycles to operations whose mode bypasses every gate. The
trick is in what the gating logic indexes against.

\paragraph{Reused counters.}
Each gate reads the same three inputs: a per-Jetty sequence
counter (``do prior operations from this application still need
to complete?''), a per-channel outstanding-window counter (``can
the wire absorb another?''), and a per-Jetty fence latch. The
state model in \S\ref{subsec:design-state} already holds all
three --- the spec defines the relevant sequence numbers as
per-Jetty objects, and the per-Jetty table carries them
regardless. The ordering surface reads from counters the state
model writes to; it does not add a parallel bookkeeping
structure.

\begin{figure*}[t]
  \centering
  \includegraphics[width=0.95\textwidth]{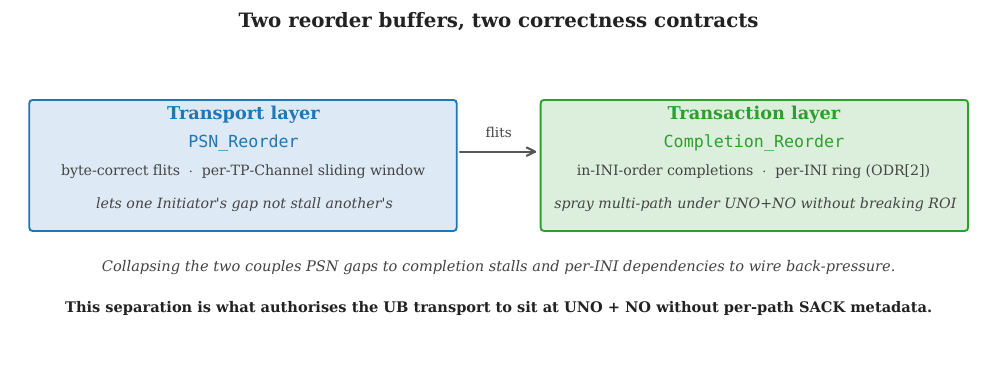}
  \caption{The pipeline carries two reorder buffers serving disjoint
  correctness contracts. Packet-sequence reordering at the transport
  layer delivers byte-correct flits regardless of application
  ordering; transaction-sequence reordering at the transaction layer
  delivers in-issue-order completions regardless of wire-byte order.
  The separation is what makes multi-path spreading and opt-in
  ordering composable.}
  \label{fig:two-reorder}
\end{figure*}

\paragraph{Two reorder buffers, not one.}
A subtler choice is to carry \emph{two} reorder buffers serving
disjoint contracts (Figure~\ref{fig:two-reorder}). One acts at
the transport layer on packet sequence numbers and delivers
byte-correct flits regardless of any application's ordering
choice, so a TP~Channel can multiplex multiple applications
without one application's wire-byte gaps stalling another's. The
other acts at the transaction layer on per-Jetty sequence numbers
and delivers in-issue-order completions regardless of wire-byte
order, so the transmit path can spread packets across multiple
network paths without breaking the application's view. Collapsing
the two would couple the contracts: every multi-path packet gap
would block the completion stream; every per-application
dependency stall would back-pressure the wire. The separation is
what makes unordered service mode and strict mode correctness-safe
at the same time. \S\ref{sec:related} contrasts this with the per-transaction
bitmap approach of the Ultra Ethernet Consortium (UEC) and
MP-RDMA.

\subsection{Realising the load/store path}
\label{subsec:design-loadstore}

The work-queue-driven path runs ten pipeline stages cold (work-queue
scheduling, ordering, header build, transport accounting,
congestion check, multi-path lane select, retransmit-buffer write,
framing, transmit). For a small synchronous operation, every one
is overhead the abstraction does not need.

The load/store path drops them. A CPU \texttt{ld} or \texttt{st}
to the controller's address aperture builds the transaction header,
attaches a transport-bypass flag, builds the network header, and
frames the packet for the wire --- five stages cold. No work-queue
entry, no completion entry, no sequence-number allocation (the
bypass flag tells the receiving NIC to skip its transport-layer
state machine), no retransmit-buffer slot (the CPU re-issues the
load on timeout, which is cheaper than transport-layer reliability
on a path that is synchronous anyway). The path admits only
operations whose semantics survive transport bypass --- small
loads, stores, and same-path atomics --- and the controller must
sit on the CPU's on-chip-bus segment; bulk and asynchronous
transfers stay on the work-queue path.

\subsection{Hardware fanout: target-side dispatch}
\label{subsec:design-fanout}

\begin{figure*}[t]
  \centering
  \includegraphics[width=0.95\textwidth]{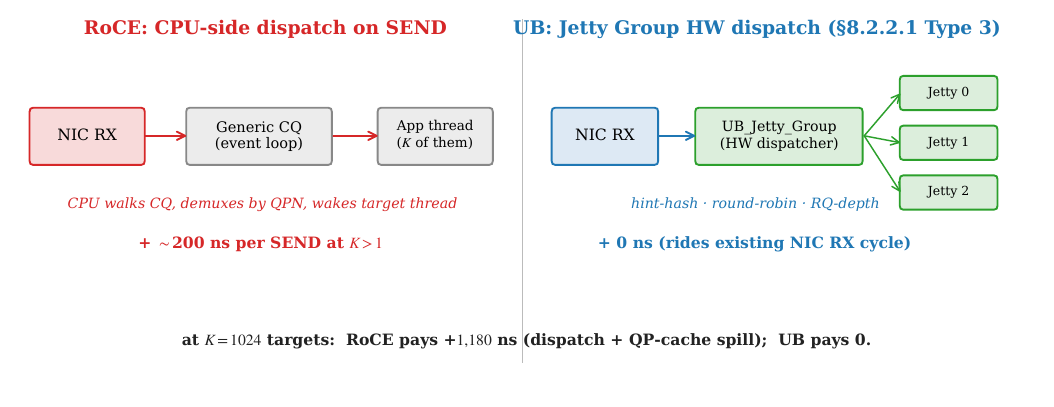}
  \caption{Target-side SEND dispatch. RoCE demultiplexes
  through a shared completion queue plus an application event loop;
  UB's Jetty~Group performs the dispatch in hardware on the
  membus crossing already paid for by NIC RX.}
  \label{fig:jetty-group-arch}
\end{figure*}

Bounded state has a second consequence: target-side
demultiplexing --- which local application receives an incoming
SEND --- can live in hardware rather than in a CPU event loop.
RoCE handles this in software (the target NIC writes a completion
to a shared queue and the application event loop dispatches at
CPU speed). UB defines a \emph{Jetty Group} in which a single
externally visible identifier represents a set of member Jetties;
the receiving NIC dispatches each incoming SEND to a member in
hardware under one of three policies (hash on an initiator-supplied
key, round-robin, or receive-queue-depth load balancing)
(Figure~\ref{fig:jetty-group-arch}). The target CPU is no longer
on the SEND fast path. The cost: one extra RX pipeline element
and $\sim$80~B per 8-member group. \S\ref{subsec:m2n} measures
the saving at ${\sim}200$~ns per SEND for small fanout, growing to
${\sim}1200$~ns once RoCE's QP spill compounds with the CPU
dispatch.

\subsection{What the implementation makes visible}
\label{subsec:design-composition}

The implementation puts numbers on the dependencies the
introduction asserted. The on-chip controller needs bounded
state: with per-Jetty and per-TP-Channel tables fitting in
$\sim$110~KB at $(1024,1024)$, every load/store reaches the
relevant context in on-chip BRAM; were the state to spill to host
DRAM, the controller would pay a host-memory access on every
operation and lose its latency edge. Opt-in ordering needs the
layer split: the gating logic indexes per-Jetty counters that the
table already provisions, costing $\sim$13~KLUT and zero pipeline
cycles on operations that bypass gating; a QP-centric design
would have to add those counters from scratch on every operation.
Multi-path spreading needs the two-reorder-buffer split: the
transport- and transaction-layer reorder buffers
(\S\ref{subsec:design-ordering}) carry disjoint contracts, so an
unordered application receives unordered completions while an
ordered application sharing the same wire receives ordered ones;
collapsing the buffers would force one contract onto both, and
the unordered mode could no longer authorise spreading. The next
sections quantify what the resulting design costs
(\S\ref{sec:impl}) and what it saves (\S\ref{sec:eval-state}
onwards).

\section{Implementation}
\label{sec:impl}

OpenURMA is $\sim$3.5~KLOC of element-source across 39 pipeline
elements spanning the transport and transaction layers; the
parallel OpenRoCE baseline (RoCEv2~RC) is $\sim$1.3~KLOC across 21
elements. Both sit on the same OpenClickNP
toolchain~\cite{openclicknp}, so every comparison is between
artifacts from an identical lowering pipeline. We describe each
piece by what it does, not what it is named.

\subsection{Pipeline decomposition}
\label{subsec:impl-elements}

The 39 elements break into six categories whose counts the
protocol fixes. \textbf{Ten header parser/builder pairs} handle
the four layered headers (Ethernet framing, network-transport,
reliable- and unreliable-transport, and base-transaction).
\textbf{Nine transport-layer elements} implement the per-host
TP~Channel pipelines, the retransmission and packet-sequence
reorder buffers, the per-packet and transaction-level
acknowledgement generators, the congestion-control window and its
echo-feedback element, and the multi-path dispatcher.
\textbf{Four ordering elements} realise the gating surface of
\S\ref{subsec:design-ordering}: the Jetty scheduler, a
completion reorder buffer, and initiator- and target-side order
trackers (two because the gating responsibility shifts with the
service mode). \textbf{Seven data-path elements} cover memory
read/write, atomics, per-Jetty receive, the target-side
dispatcher, the opcode router, and the completion generator. \textbf{Three state tables} hold
memory-region permissions, per-Jetty state, and per-TP-Channel
state. \textbf{Six glue elements} provide the doorbell, dispatch
multiplexers, completion-notification tees, completion-queue
streamers, and the retransmission RTO timer.

Two decompositions carry design intent: the initiator- and
target-side order trackers are separate because gating shifts
between endpoints with the service mode (relaxed at the target,
strict at the initiator), and the completion reorder buffer is
distinct from the packet-sequence reorder buffer because the two
serve disjoint correctness contracts on disjoint sequence-number
spaces (\S\ref{subsec:design-ordering}). The work-queue transmit
path threads ten stages cold, doorbell to Ethernet encapsulation;
the receive path inverts it (Figure~\ref{fig:topology}).

Two elements are new. The \emph{load/store bypass engine}
(\S\ref{subsec:design-loadstore}) is a separate topology variant
that takes a load/store doorbell off the on-chip bus, allocates a
one-shot context (no sequence number, no retransmission slot),
stamps a transport-bypass flag, and frames the packet --- its first
wire flit emerges at \textbf{8 cycles ($\approx$25~ns)}, 16 below
the work-queue path. The \emph{target-side dispatcher}
(\S\ref{subsec:design-fanout}) rewrites a SEND addressed to a
Jetty-Group identifier to a member Jetty (hint-hash, round-robin,
or queue-depth balancing; unregistered identifiers pass through),
at $\sim$80~B per 8-member group.

\subsection{Retransmission and congestion control}
\label{subsec:impl-retrans}

The retransmission buffer is a 64-slot per-TP-Channel ring
supporting both Go-Back-N and selective retransmit; the selective
path uses the spec's selective-ack opcode and a
per-packet-sequence-number (PSN) bitmap, replayed on an explicit
NAK or the RTO timer. Each slot holds one metadata/extension flit
pair --- enough for control-plane ops and $\le$8-byte Writes;
multi-flit Write replay (a payload-flit list per slot) runs on the
loss-free path but is not yet in the buffer
(\S\ref{sec:discussion}). Congestion control pairs a host-side
switch model (queue watermarks stamping the FECN bit) with a
per-channel additive-increase / multiplicative-decrease (AIMD)
window; in a 200-WR closed-loop test, 9 of 25 packets are marked
and the sender's window backs off from 65{,}536~B to 4{,}096~B.

\subsection{Toolchain and back-ends}
\label{subsec:impl-backends}

OpenClickNP lowers each element from one source through three
back-ends: a thread-and-FIFO software emulator (correctness tests,
\S\ref{sec:correctness}); a cycle-accurate SystemC simulator on a
1~ns clock matching the 322~MHz target, so cycle counts compare
directly to post-route timing (\S\ref{sec:sc}); and a Vitis HLS
back-end that Vivado places and routes (\S\ref{sec:vivado}). An
element that closes timing in HLS is exactly the one the emulator
tested and the simulator benchmarked. Our only framework change
was two element-source pragmas --- one to declare an element's
initiation interval explicitly, one to forward arbitrary HLS
pragmas --- needed because the back-end previously hard-coded
$II{=}1$, over-pipelining six elements whose data dependencies
cannot be unrolled. Both are upstream and reused by both stacks.

\subsection{Two-node SystemC simulator}
\label{subsec:impl-twonode}

The two-node simulator links each NIC stack as a static SystemC
library and connects two instances through an EtherLink module with
configurable bandwidth and delay. Each endpoint is a full system
--- host CPU posting doorbells, the NIC TX/RX pipelines, a DDR4
model, a write-back / write-through / uncached cache hierarchy, and
the completion-queue write-back path --- and the harness sweeps 388
configurations across the six verb categories, four cache policies,
8~B--4~KB payloads, and 100~ns--10~$\mu$s link delays.

\subsection{gem5 full-system scaffold}
\label{subsec:impl-gem5}

For the CPU-to-remote-DRAM end-to-end claim we embed the SystemC
NIC into gem5~\cite{lowepower2020:gem5} as a memory-mapped device,
so an ARM CPU running Linux~4.14, the \texttt{uburma.ko} driver,
and a libc-linked benchmark drive the cycle-accurate pipeline
through a real OS stack rather than a synthetic injector. The
integration took three non-obvious steps --- rebuilding the SC
libraries against gem5's embedded (ABI-incompatible) SystemC,
carving the NIC aperture out of system memory so gem5's crossbar
can route it, and draining the TLM pipeline synchronously on each
transaction since the atomic CPU never yields to the SC event
queue --- but none touch the protocol. Bringing the full
WRITE~$\rightarrow$~TAACK (transaction-layer acknowledgement)
~$\rightarrow$~CQE roundtrip up end-to-end also surfaced three
latent pipeline bugs the synthetic-injector tests had masked: a
dropped service-mode field, a stranded reorder-buffer flit, and an
unrouted received-ACK port. Fixing them is what lets the scaffold
report the three completion-queue-entry (CQE) delivery paths of
\S\ref{sec:methodology}.

\subsection{Timing-closure sweep}
\label{subsec:impl-closure}

The initial Vitis-HLS pass on the 38 work-queue-driven elements
left 8 needing remediation: 5 missed 322~MHz in Vivado P\&R and 3
failed at HLS before reaching P\&R. The fixes combined the new
$II$/pragma extensions with a few algorithmic redesigns
(Table~\ref{tab:timing-closure}).

\begin{table}[h]
  \centering
  \scriptsize
  \setlength{\tabcolsep}{3pt}
  \begin{tabular}{@{}lll@{}}
  \toprule
  Element function & Before $\to$ After (ns WNS) & Action \\
  \midrule
  Atomic operations           & $-1.871 \to +0.298$ & $II{=}4$ + mem partition \\
  Completion reorder          & HLS-stuck $\to +0.672$ & head-ring + $II{=}2$ \\
  Initiator order tracker     & $-5.382 \to +0.318$ & $II{=}2$ \\
  Jetty scheduler             & HLS-aborted $\to +0.546$ & $II{=}2$ \\
  Target order tracker        & $-2.25 \to +0.101$ & $II{=}2$ + drain reorder \\
  On-NIC memory write         & $-0.628 \to +0.628$ & memory partition \\
  Memory-region table         & failing $\to +0.729$ & shrink to 64 entries \\
  On-NIC memory read          & $-0.449 \to +0.282$ & word-sized array \\
  \bottomrule
  \end{tabular}
  \caption{Timing-closure remediation, by element function.}
  \label{tab:timing-closure}
\end{table}

\paragraph{The instructive case.}
The on-NIC memory-read element first backed its 64~KB path with a
byte-array plus cyclic partition for parallel byte-bank access. HLS
estimated 490~MHz, but Vivado P\&R missed by $0.449$~ns: the
critical path was 8~LUT levels of barrel-shift mux into the
partitioned BRAM, routing-bound, and no $II$ or partition factor
moved it. The fix was at the data-layout level --- an
8-byte-word array indexed by a shifted offset turns each aligned
read into one BRAM word access, eliminating the mux and closing at
$+0.282$~ns (the same change applied to OpenRoCE). The lesson: HLS's
per-element timing estimate is optimistic, routing-bound paths only
appear in P\&R, and the fix is often in data layout, not pragmas.

\section{Evaluation methodology}
\label{sec:methodology}

The evaluation is organised around the three design commitments.
\S\ref{sec:feasibility} establishes that the NIC is synthesizable,
cheap, correct, and faithfully modeled;
\S\ref{sec:eval-state}--\S\ref{sec:eval-ordering} validate each
commitment in turn (bounded state, the load/store latency collapse,
and near-free opt-in ordering); \S\ref{sec:eval-fullsystem} confirms
all three under a real OS; and \S\ref{sec:eval-transport} and
\S\ref{sec:eval-scaleout} cover reliable transport and the scale-out
reach that coherent fabrics cannot match. This section first
describes the measurement substrate the rest of the evaluation
shares.

\subsection{Three modeling tiers and a matched baseline}
\label{subsec:tiers}
Without physical silicon, every number comes from one of three
back-ends lowered from the \emph{same} element source
(\S\ref{subsec:impl-backends}), so a kernel that passes correctness
is the one that is benchmarked and synthesised:
\textbf{(1)} a thread-and-FIFO software emulator answers
correctness questions; \textbf{(2)} a cycle-accurate SystemC
simulator (1~ns clock matching 322~MHz) reports latency,
throughput, and ordering cost, either standalone per element or
wired into a two-node model whose analytical surrounds (host CPU,
PCIe/membus, wire, DRAM) close the end-to-end path; and
\textbf{(3)} a gem5 full-system scaffold runs Linux + the uburma
driver + a real benchmark against the SystemC NIC. A matched
\textbf{OpenRoCE} (RoCEv2~RC) baseline is lowered through the same
toolchain at every tier, so all comparisons hold every variable
below the protocol fixed. Every figure is reproducible from one
CSV-emitting harness.

Four \emph{configurations} appear as columns throughout:
\textbf{UB LD/ST} (the \specsec{8.3} load/store transport-bypass topology),
\textbf{UB URMA} (the \specsec{8.4} work-request path), and \textbf{RoCE
BF} / \textbf{RoCE DMA} (the OpenRoCE stack with Blue-Flame
inline-WQE or standard DMA-fetched WQE). BF and DMA are runtime
modes of one RoCE stack; LD/ST and URMA are two topologies of the
OpenURMA pipeline. NIC TX/RX cycle counts come from the
microbenchmarks (\S\ref{sec:feasibility}): 8 for UB LD/ST, 9 for
RoCE RC, and 25 for UB URMA --- one cycle above its measured
24-cycle cold path, a rounding that charges UB slightly more
latency than it shows.

\subsection{Bidirectional cost model}
\label{subsec:costmodel}
\begin{figure*}[t]
  \centering
  \includegraphics[width=0.95\textwidth]{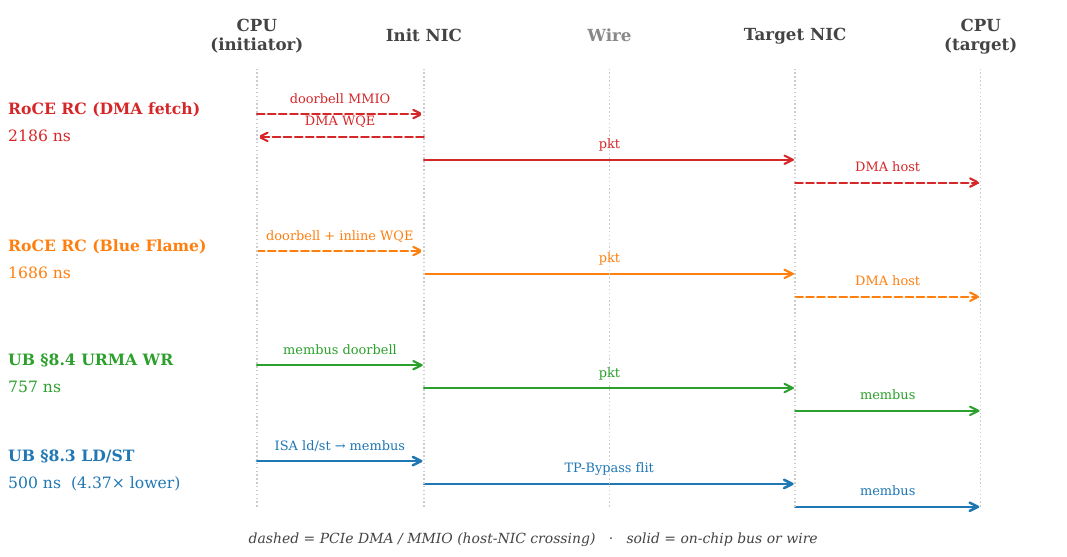}
  \caption{Submission path per stack. RoCE traversals between CPU and
  NIC (dashed) sit on PCIe; UB carries the same hand-offs over the
  on-chip bus. On UB LD/ST the verb \emph{is} an ISA instruction,
  and the wire crossing carries a TP-Bypass flit instead of a full
  RTPH-wrapped packet.}
  \label{fig:submission-path}
\end{figure*}

The single-node RDMA-latency literature routinely collapses
submission and completion into black boxes and treats the target as
a free ``remote DRAM hit'' --- hiding exactly the costs UB removes.
We instead make every host$\leftrightarrow$NIC interaction on
\emph{both} sides of the wire an explicit critical-path component:
the target pays its own RX/TX pipeline cycles, its own
NIC$\leftrightarrow$DRAM transfer, and the DRAM row hit before it
can build the response. Figure~\ref{fig:submission-path} sketches
each round trip (PCIe dashed, on-chip/wire solid). The accounting
principle is that every traversal a single-node treatment folds
away --- the doorbell, the WQE fetch, the initiator-side response
DMA, the CQE write, and above all the \textbf{Target
NIC$\leftrightarrow$DRAM} transfer hidden inside the ``remote DRAM
hit'' --- is charged explicitly on the critical path, for both
stacks and both directions. \S\ref{subsec:lat-headline} populates
this decomposition row by row for a 64~B READ
(Table~\ref{tab:per-side}), reports the headline totals, and reads
off where each stack's round trip goes; here we fix only the
accounting and the parameters it runs on.

Parameter defaults follow published ConnectX-7-class
ranges~\cite{ramos2023:multicluster,kaufmann2020:tas}, each a
command-line knob. A two-level cache + LLC + local-DRAM hierarchy
(fully-associative LRU; WB/WT/UC; 1/4/12/70~ns hit latencies) is
consulted on every \specsec{8.3} \texttt{ld}/\texttt{st}; WR and RoCE verbs
target the wire and bypass it. The wire defaults to 100~ns at
400~Gbps and remote DRAM to a 30~ns row hit. We model polling
completion only --- the regime high-performance RDMA uses --- and
omit event-driven completion, whose MSI-X / interrupt / scheduler /
wakeup costs are OS-dependent and irreducible to one parameter
(FastWake~\cite{li2023:fastwake} characterises that path; the gem5
tier measures it directly, \S\ref{sec:eval-fullsystem}). The harness
sweeps all six verb categories across four link delays, four
in-flight depths, 8~B--64~KB payloads, three cache policies, and
four locality points --- 388 valid configurations.

\subsection{Modeling caveats}
\label{subsec:twonode-caveats}
Five caveats, in decreasing order of weight. \textbf{(i)} NIC cycle
counts (8/25/9) are measured, not modeled --- the UB URMA value
rounded conservatively upward. \textbf{(ii)} The surrounding
analytical parameters track published ConnectX-7-class ranges
rather than a specific target; each is a knob, and the link-delay
sweep (\S\ref{sec:eval-latency}) shows the qualitative ordering is
robust. \textbf{(iii)} The cache model is fully-associative LRU,
consulted only for \specsec{8.3} verbs as the spec requires.
\textbf{(iv)} Completion is polling-only in SystemC; the gem5 tier
covers event-driven completion. \textbf{(v)} No CPU
microarchitecture is modeled in SystemC --- it reports the
controller-to-controller floor, against which any CPU model adds
latency, never removes it; the gem5 atomic-CPU run validates this
monotonicity, and an out-of-order ARM config is exposed as a knob.

\section{Feasibility and validation}
\label{sec:feasibility}

Before measuring the design's payoff, we establish that the NIC is
real, cheap, correct, and faithfully modeled: every element
synthesises and closes 322~MHz on the Alveo~U50 (\S\ref{sec:vivado}),
the area cost is a small fraction of the device, the raw
NIC-pipeline latency is characterised cycle-by-cycle, the functional
suite passes (\S\ref{sec:correctness}), and the analytical baseline
reproduces published ConnectX-7 silicon within $\pm$5\%
(\S\ref{subsec:c1-validation}).

\subsection{The pipeline synthesises and closes timing}
\label{sec:vivado}
We run Vitis HLS C-synthesis on every element, then drive Vivado
2025.2 through out-of-context place-and-route per element at the
U50's 3.106~ns / 322~MHz target. \textbf{All 38 work-queue-driven
OpenURMA elements and all 21 OpenRoCE elements close 322~MHz
post-route} with positive worst negative slack
($+$0.079~\dots~$+$1.425~ns; the load/store bypass element
synthesises into the same header-build chain and is not a separate
row).

\begin{table}[h]
  \centering
  \scriptsize
  \setlength{\tabcolsep}{4pt}
  \begin{tabular}{lrrr}
  \toprule
  Metric & OpenURMA (38) & OpenRoCE (21) & Ratio \\
  \midrule
  LUT & 122{,}710 & 46{,}636 & $2.63\times$ \\
  FF & 194{,}266 & 91{,}900 & $2.11\times$ \\
  BRAM18 & 328 & 67 & $4.90\times$ \\
  DSP & 3 & 0 & --- \\
  Meeting 322~MHz & \textbf{38 / 38} & \textbf{21 / 21} & --- \\
  WNS min (ns) & $+0.079$ & $+0.103$ & --- \\
  WNS max (ns) & $+1.425$ & $+1.394$ & --- \\
  \bottomrule
  \end{tabular}
  \caption{Aggregate post-route resource and timing summary for the
    two stacks on the Alveo~U50 at the 322~MHz target.}
  \label{tab:resource-summary}
\end{table}

OpenURMA fits in 14.1\% of the U50's LUT budget, 11.1\% of
flip-flops, and 12.2\% of BRAM18 (OpenRoCE: 5.4 / 5.3 / 2.5\%),
leaving ample room for the platform shell ($\sim$50--80K LUTs of
fixed MAC/DMA/NoC overhead). The 8 elements that initially missed
timing (5 in P\&R, 3 at HLS) closed with the $II$/pragma
extensions plus a few data-layout redesigns
(\S\ref{subsec:impl-closure}).

\subsection{Where the area goes}
\begin{figure}[t]
  \centering
  \includegraphics[width=\columnwidth]{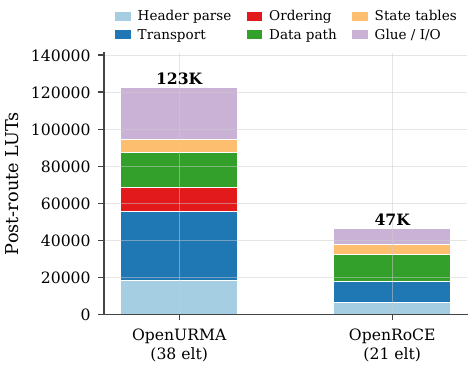}
  \caption{Post-route LUT budget by architectural role for both
  stacks.}
  \label{fig:vivado_breakdown}
\end{figure}

Figure~\ref{fig:vivado_breakdown} categorises every element by
architectural role.

\begin{table}[h]
  \centering
  \footnotesize
  \begin{tabular}{lrrr}
  \toprule
  Category (LUTs) & OpenURMA & OpenRoCE & $\Delta$ \\
  \midrule
  Header parse/build         & 18,394 & 6,657   & +11,737 \\
  Transport (reliable)       & 37,103 & 11,094  & +26,009 \\
  Ordering / completion      & 13,036 & ---     & +13,036 \\
  Data path                  & 18,806 & 14,822  & +3,984  \\
  State tables               &  6,880 &  4,997  & +1,883  \\
  Glue / I/O                 & 28,491 &  9,066  & +19,425 \\
  \midrule
  Total                      & 122,710 & 46,636 & +76,074 \\
  \bottomrule
  \end{tabular}
  \caption{Post-route LUT budget by architectural category for the
    two stacks (the per-role data visualised in
    Figure~\ref{fig:vivado_breakdown}).}
  \label{tab:area-category}
\end{table}

\begin{figure}[t]
  \centering
  \includegraphics[width=\columnwidth]{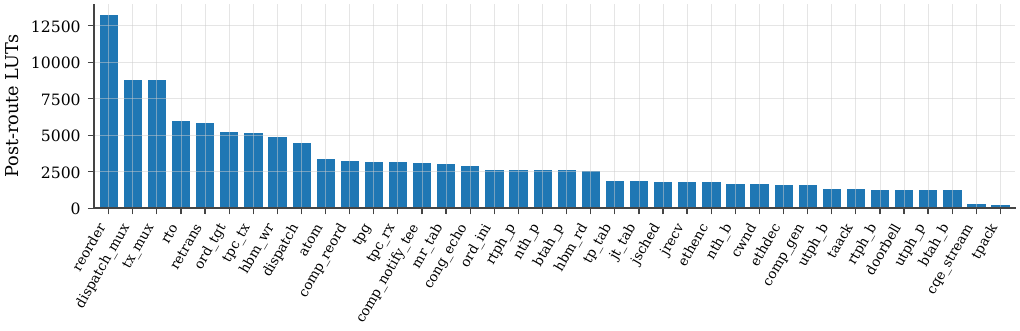}
  \caption{Per-element post-route LUT, sorted descending. All 38
  OpenURMA elements meet 322~MHz with positive WNS; the state-heavy
  elements dominate the budget.}
  \label{fig:per_elt_lut}
\end{figure}

OpenURMA's $\sim$76~KLUT excess over OpenRoCE breaks down as
$\sim$13~KLUT for the opt-in ordering surface (the four ordering
elements); $\sim$26~KLUT for the richer transport layer
(selective-retransmit ring, packet-sequence reorder buffer,
congestion-echo, multi-path spreading --- vs RoCE's plain Go-Back-N
and DCQCN); $\sim$12~KLUT for splitting the network-, reliable-,
and unreliable-transport headers into separate parser/builder pairs
(vs RoCE's single combined header); $\sim$4~KLUT for the data path
(the full atomic opcode set, a superset of RoCE's); and
$\sim$19~KLUT of glue. Within glue, the two 4-input round-robin
muxes (dispatch and transmit) cost 8.8~KLUT each --- 14.4\% of the
total --- versus 3.6~KLUT in OpenRoCE; the 2.4$\times$ expansion is
the wider flit lanes UB carries (three protocol headers vs one), an
HLS-instantiation artifact a single wider crossbar would close on
tape-out. Figure~\ref{fig:per_elt_lut} sorts per-element LUT; the
head is the state-heavy elements (retransmission buffer, the two
reorder buffers, target-side order tracker, TP-Channel transmit).

\subsection{Raw NIC-pipeline latency}
\label{sec:sc}
\begin{figure*}[t]
  \centering
  \begin{subfigure}[t]{0.49\textwidth}
    \includegraphics[width=\linewidth]{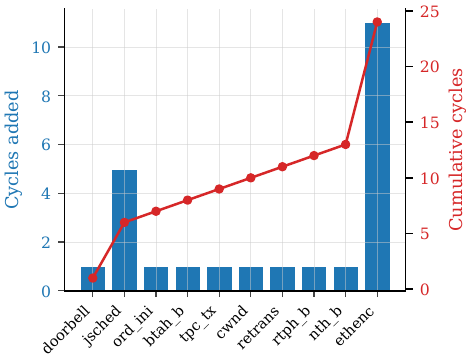}
    \caption{Per-stage cycle cost (cumulative 24~cy at the wire).}
    \label{fig:per_element_latency_a}
  \end{subfigure}
  \begin{subfigure}[t]{0.49\textwidth}
    \includegraphics[width=\linewidth]{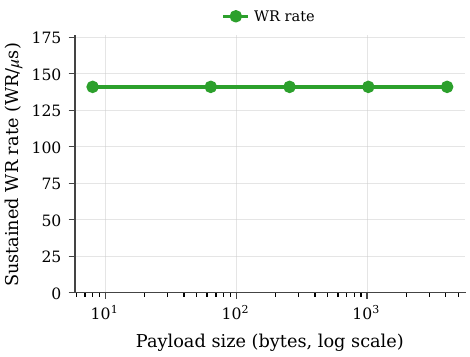}
    \caption{Sustained WR rate vs payload --- header-rate-limited.}
    \label{fig:throughput}
  \end{subfigure}
  \caption{Raw NIC-pipeline microbenchmarks. (a) Per-stage cycle
  contribution, cumulative 24~cy at the wire. (b) Sustained WR rate
  vs payload: header-rate-limited in the small-to-medium regime.}
  \label{fig:sc_micro}
\end{figure*}

A single Write WR (ROL+NO) drives the TX pipeline cold; tap monitors
between adjacent stages record first-flit arrival
(Fig.~\ref{fig:per_element_latency_a}). The slowest stages are the
Jetty scheduler (5 cycles --- $II{=}2$ plus the round-robin scan)
and Ethernet encapsulation (11 cycles --- byte-stream-rate); the
initiator-side order tracker costs 1 cycle on the unblocked path.
Total cold-path TX latency is \textbf{24 cycles ($\approx$75~ns at
322~MHz)}. Sweeping payload from 8~B to 4~KB
(Fig.~\ref{fig:throughput}), the per-WR rate is constant at
$\approx$141~WR/$\mu$s: the pipeline is metadata-flit-rate-limited
in the small-to-medium regime that dominates AI-collective control
and HPC small-message traffic, and at 4~KB (129 flits/WR) the same
rate is $\approx$4.6~Tb/s of wire bytes --- bandwidth is not the
binding constraint at any payload size. This 24-cycle floor and the
header-rate limit are the substrate against which the load/store
collapse (\S\ref{sec:eval-latency}) and the free ordering surface
(\S\ref{sec:eval-ordering}) are measured.

\subsection{Functional correctness}
\label{sec:correctness}
A 17-test software-emulator suite covers the spec surface in four
groups, all passing. \textbf{Ordering} exercises the full \specsec{7.3}
surface: initiator- and target-side gating, Fence, in-issue versus
arrival-order completion, fused acknowledgement, unordered Send,
mixed-mode queues, per-initiator parallelism, and
head-of-line-blocking isolation. \textbf{Data path} covers on-NIC
read/write integrity, the full atomic opcode set (swap, load,
store, fetch-and-\{add, sub, and, or, xor\}, each returning the
pre-modification value with compare-and-swap under the mixed-mode
test), and the multi-flit Write payload path (8--256~B through
encap/decap). \textbf{Wire format} checks the encoder against the
spec bit layout and a full-header transmit round trip.
\textbf{System-level} covers a 50{,}000-WR throughput run, the
end-to-end congestion loop (\S\ref{sec:eval-transport}), and
target-side hardware dispatch across all three Jetty-Group policies
(\S\ref{sec:eval-state}). The one gap is loss-path replay of
\emph{multi-flit} Writes: the retransmit slot stores one
(meta,~ext) pair, so a dropped multi-flit Write does not yet
re-issue its payload (a follow-on item, \S\ref{sec:discussion});
all single-flit retransmit paths are exercised.

\subsection{Model validation against ConnectX-7}
\label{subsec:c1-validation}
\begin{figure}[t]
  \centering
  \includegraphics[width=0.95\columnwidth]{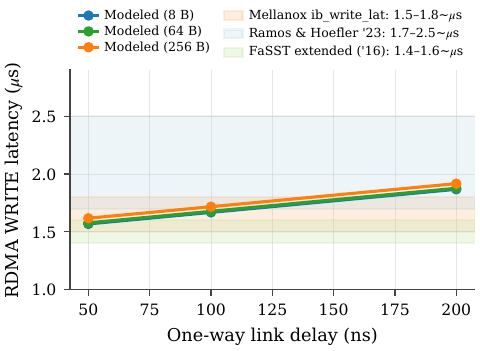}
  \caption{Modeled RoCE-DMA RDMA WRITE latency (curves) vs published
  ConnectX-7 ranges (bands).}
  \label{fig:c1-validation}
\end{figure}

We anchor the analytical defaults against published silicon. The
simulator's RoCE-DMA stack yields 1.57--1.62~$\mu$s for an 8~B RDMA
WRITE at 50~ns link delay and 1.67--1.72~$\mu$s at 100~ns
(Figure~\ref{fig:c1-validation}), within $\pm$5\% of Mellanox's
ConnectX-7 measurements (1.5--1.8~$\mu$s,
\cite{ramos2023:multicluster,kaufmann2020:tas}) and FaSST's
extended numbers (1.4--1.6~$\mu$s, \cite{kalia:fasst-extended}),
supporting every downstream RoCE-vs-UB ratio. These published
ranges play two distinct roles in the paper, which we keep separate:
\emph{here} they validate that our RoCE model reproduces real
silicon, calibrating the baseline; \emph{later}
(\S\ref{sec:eval-fullsystem}) the same ranges serve as an external
comparison target for the gem5 full-system stack, reporting a result
against it.

\section{Bounded state scales additively}
\label{sec:eval-state}

The first commitment is that splitting per-application endpoint
state (Jetty) from per-host transport state (TP~Channel) makes
per-NIC state grow as $O(N{+}M)$ rather than RoCE's
$O(N{\cdot}M)$. We validate this from the byte count down to the
end-to-end latency cliff it removes.

\subsection{Per-NIC state, field by field}
\begin{figure}[t]
  \centering
  \includegraphics[width=\columnwidth]{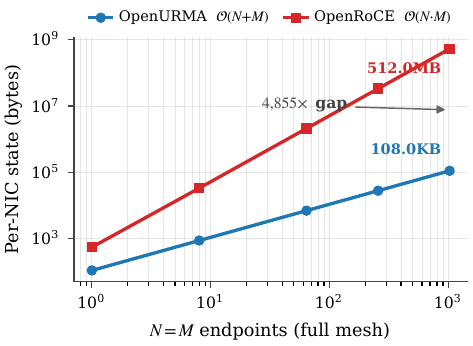}
  \caption{Per-NIC connection state vs endpoint count $(N{=}M)$:
  OpenURMA's $O(N{+}M)$ vs RoCE's $O(N{\cdot}M)$, reaching
  4{,}855$\times$ at 1024 endpoints.}
  \label{fig:state}
\end{figure}

Enumerating the actual state-struct fields each element holds
(Table~\ref{tab:state-decomp}), a per-Jetty record is 20~B, a
per-TP-Channel record 56~B, and the per-application memory-region
record 32~B. OpenURMA's total is the sum of three additive tables
($N$ Jetties, $N$ MR records, $M$ TP~Channels) --- 110.6~KB at
$(1024,1024)$ --- while RoCE's $N{\cdot}M$ QP pool dominates its
537~MB. The asymptotic ratio reaches \textbf{4{,}855$\times$}
(Table~\ref{tab:state-ratio}; Fig.~\ref{fig:state}). Projecting the
full \specsec{8.2.2} spec surface (suspend-mode drain, exception counters,
public-Jetty owner, Jetty-Group back-pointer) grows the Jetty to
48~B and softens the ratio only to 3{,}855$\times$ --- the residual
spec fields the MVP elides do \emph{not} explain the gap; the
$(N{+}M)$ vs $(N{\cdot}M)$ split does. At $(1024,1024)$ that gap
straddles the boundary between fits-in-on-chip-SRAM and
spill-to-host-DRAM for a typical NIC.

\begin{table}[h]
  \centering
  \footnotesize
  \setlength{\tabcolsep}{4pt}
  \caption{Per-connection state, decomposed against
    UB-Base-Specification 2.0.1 fields. MVP = today's per-Jetty
    record; full-spec adds state-machine drain, exception-mode,
    public-Jetty owner, and Jetty-Group back-pointer. Even the
    full-spec descriptor stays under 10\% of RoCE's per-QP context.}
  \label{tab:state-decomp}
  \begin{tabular}{lr|lr|lr}
  \toprule
  Jetty (MVP) & B & Jetty (full \specsec{8.2.2}) & B & TP Channel & B \\
  \midrule
  jetty\_id        & 4  & + sq/rq handle    & 8  & remote\_cna   & 4  \\
  token\_value     & 4  & + jfae\_id        & 4  & local/rem tpn & 8  \\
  jfc\_id          & 4  & + public/owner    & 4  & psn\_next     & 4  \\
  type             & 1  & + drain bookkeep. & 6  & tpmsn\_next   & 4  \\
  state            & 1  & + exc.\ mode      & 1  & last\_acked   & 4  \\
  valid            & 1  & + fault counter   & 2  & flags         & 2  \\
  align pad        & 5  & + mr\_perm idx    & 2  & epsn (RX)     & 4  \\
                   &    & + group back-ptr  & 4  & emsn (RX)     & 4  \\
                   &    & + (base 20 B)     & 20 & base\_psn     & 4  \\
                   &    & + align           & -3 & sack\_bitmap  & 8  \\
                   &    &                   &    & max\_rcv\_psn & 4  \\
                   &    &                   &    & mode flags    & 2  \\
                   &    &                   &    & align pad     & 4  \\
  \midrule
  \textbf{Total}   & \textbf{20} & \textbf{Total} & \textbf{48} & \textbf{Total} & \textbf{56} \\
  \bottomrule
  \end{tabular}
\end{table}

\begin{table}[h]
  \centering
  \scriptsize
  \setlength{\tabcolsep}{3pt}
  \caption{Per-NIC connection state vs endpoint count.}
  \label{tab:state-ratio}
  \begin{tabular}{@{}lrrr@{}}
  \toprule
  $(N, M)$ & OpenURMA & OpenRoCE & Ratio \\
  \midrule
  $(1, 1)$ & 108~B & 544~B & $5.0\times$ \\
  $(8, 8)$ & 864~B & 33~KB & $38\times$ \\
  $(64, 64)$ & 6.9~KB & 2.1~MB & $304\times$ \\
  $(256, 256)$ & 27.6~KB & 33.6~MB & $1{,}214\times$ \\
  $\mathbf{(1024, 1024)}$ & $\mathbf{110.6\,KB}$ & $\mathbf{536.9\,MB}$ & $\mathbf{4{,}855\times}$ \\
  $(1024, 1024)$ full-spec & 139.3~KB & 536.9~MB & $3{,}855\times$ \\
  \bottomrule
  \end{tabular}
\end{table}

\subsection{The byte ratio becomes a latency cliff}
\label{subsec:sram-spill}
\begin{figure}[t]
  \centering
  \includegraphics[width=\columnwidth]{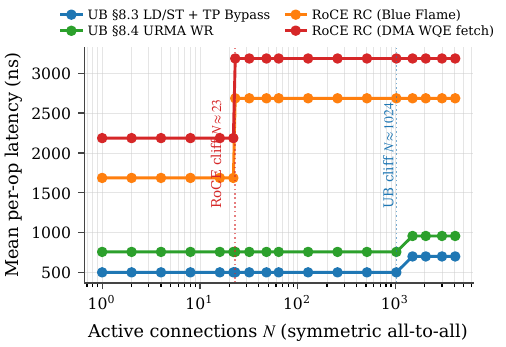}
  \caption{NIC SRAM context-cache spill. RoCE hits the cliff at
  $N{\approx}23$ ($N^2$ QPs $>$ cache), adding $\sim$1000~ns to
  every READ; UB hits it $\sim$45$\times$ later at
  $N{\approx}1024$, and its penalty is $\sim$200~ns (membus, not
  PCIe).}
  \label{fig:sram-spill}
\end{figure}

The byte count matters because production NICs cache per-connection
context in $\sim$256~KB of SRAM; once cardinality exceeds the cache,
every operation pays a context refetch. RoCE's 512~B QPs spill at
$N{\approx}\sqrt{512}{\approx}23$; UB's 56~B TP~Channels spill an
order of magnitude later. Sweeping active connections on a 64~B
READ (Fig.~\ref{fig:sram-spill}), RoCE's per-op latency steps up by
$\sim$1000~ns at $N{\approx}23$ (a PCIe DMA refetch on both
initiator and target), while UB holds flat to $N{\approx}1024$ and
its eventual penalty is only $\sim$200~ns (membus + local DRAM, not
PCIe). Across the $N{\in}[23,1024]$ band --- exactly the operating
range of AI-training and HPC all-to-all --- RoCE pays the spill on
every operation and UB does not. This cliff is also the empirical
face of the first link in the enabling chain (\S\ref{sec:intro}): an
on-bus controller stays viable only while its state fits on-chip, so
keeping the controller on-bus \emph{without} the layer split ---
RoCE's $O(N{\cdot}M)$ regime --- reintroduces exactly the
per-operation refetch that on-bus placement exists to avoid. Bounded
state is thus load-bearing for the load/store latency of
\S\ref{sec:eval-latency}, not a separate win.

\subsection{Many-to-many fanout}
\label{subsec:m2n}
\begin{figure}[t]
  \centering
  \includegraphics[width=0.95\columnwidth]{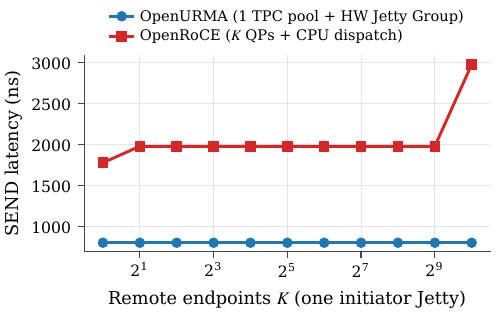}
  \caption{M2N fanout: one initiator Jetty addressing $K$ remote
  endpoints. UB flat at 804~ns (one TPC pool + HW Jetty~Group);
  RoCE jumps at $K{\geq}2$ (CPU dispatch) and again at $K{=}1024$
  (NIC SRAM spill). SEND, 64~B, link 100~ns.}
  \label{fig:m2n}
\end{figure}

For the single-initiator, $K$-target pattern of parameter-server
fan-in, UB carries all $K$ targets through one TP~Channel pool plus
$K$ cheap Jetty descriptors, and the target NIC's Jetty~Group
(\specsec{8.2.2.1} Type~3) dispatches incoming SENDs to the right member in
hardware; RoCE needs $K$ QPs and a 200~ns CPU event-loop dispatch
per target. Sweeping $K$ from 1 to 1024 (Fig.~\ref{fig:m2n}), UB
stays flat at \textbf{804~ns}; RoCE jumps from 1781~ns to 1981~ns
at $K{\geq}2$ (CPU dispatch kicks in) and to 2981~ns at $K{=}1024$
(QP-cache spill). The gap widens from 2.2$\times$ to 3.7$\times$,
all of it the layer split: the $O(N{+}M{+}K)$ cardinality keeps UB
in-cache, and the hardware Jetty~Group elides the target-side CPU
traversal.

\subsection{TP-Channel sharing}
\begin{figure}[t]
  \centering
  \includegraphics[width=\columnwidth]{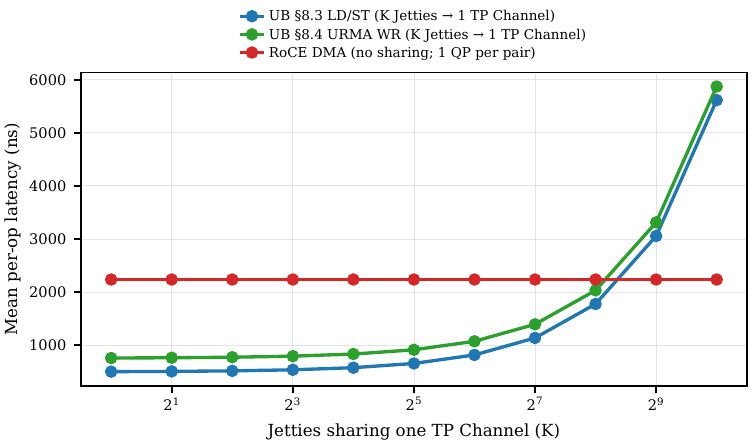}
  \caption{Per-op latency under $K$-Jetty contention on one
  TP~Channel. Linear PSN-allocator scaling; UB beats per-QP RoCE
  until $K{\approx}255$.}
  \label{fig:tp-share}
\end{figure}

A single TP~Channel can multiplex $K$ Jetties' traffic to one
remote host, each WR serialising at the channel's PSN allocator
($\sim$5~ns). Sweeping $K$ (Fig.~\ref{fig:tp-share}), UB's per-op
latency grows linearly and still beats per-QP RoCE out to
$K{\approx}255$; beyond that the allocator becomes the bottleneck,
so production should pool a small TP~Channel set per remote host
rather than share one across thousands of Jetties.

\subsection{Cluster and connection-setup scaling}
\begin{figure}[t]
  \centering
  \includegraphics[width=\columnwidth]{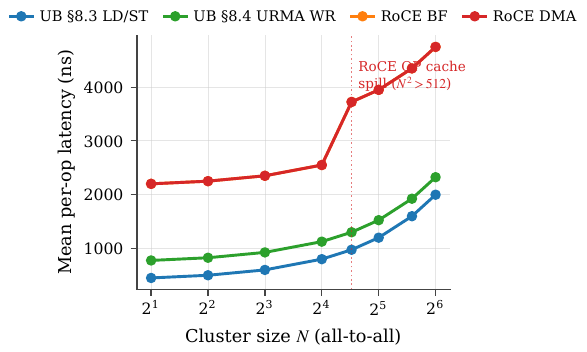}
  \caption{Cluster-scale per-op latency vs node count. RoCE crosses
  the QP-cache spill at $N{=}23$; UB stays bounded.}
  \label{fig:multinode}
\end{figure}

\begin{figure}[t]
  \centering
  \includegraphics[width=0.95\columnwidth]{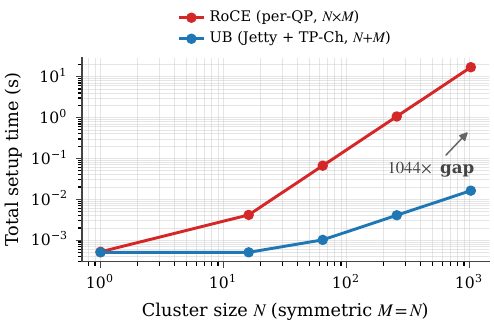}
  \caption{Total connection-setup time at symmetric $N{=}M$:
  RoCE's $O(N{\cdot}M)$ per-QP exchange (17.04~s at 1024) vs UB's
  $O(N{+}M)$ (0.016~s), a 1044$\times$ gap.}
  \label{fig:conn-setup}
\end{figure}

At cluster scale (all-to-all mesh, Fig.~\ref{fig:multinode}) UB
grows gently from 447~ns ($N{=}2$) to 1997~ns ($N{=}64$) under
wire-share contention, while RoCE starts at 2199~ns and crosses its
QP-cache cliff at $N{=}23$, reaching 4749~ns at $N{=}64$. The
control plane scales the same way: bringing up $N{=}M{=}1024$
applications costs RoCE four ioctls plus a per-QP out-of-band RTT
on each of the $N{\cdot}M$ pairs --- \textbf{17.04~s} --- versus
\textbf{0.016~s} for UB's $O(N{+}M)$ Jetty and TP~Channel setup, a
\textbf{1044$\times$} advantage (Fig.~\ref{fig:conn-setup}) that
compounds across job launches and is bounded below by the
irreducible per-QP wire RTT even with optimistic batching.

\subsection{Bounded state holds under a real OS}
The gem5 full-system tier confirms the property end-to-end: forking
$N$ tenants that each post 16 WRITEs against one NIC, the per-tenant
mean stays flat from 487~ns solo to 483~ns at $N{=}256$ (0.2\%
delta, all hits). Per-NIC SystemC state is unchanged across the run
--- only per-process Linux bookkeeping grows --- the empirical face
of the $O(N{+}M)$ claim under a real driver and scheduler.

\section{The load/store path collapses latency}
\label{sec:eval-latency}

The second commitment --- placing the controller on the on-chip bus
so a CPU load/store reaches remote memory without a PCIe traversal
--- is the paper's headline latency result. We build it up in three
movements: the headline round trip and its scaling
(\S\ref{subsec:lat-headline}); breadth across verbs, memory access,
and contention (\S\ref{subsec:lat-breadth}); and the throughput,
operating envelope, and an application port
(\S\ref{subsec:lat-app}). Confirmation under a real OS is collected
with the rest of the full-system results in
\S\ref{sec:eval-fullsystem}.

\subsection{Headline round trip and its scaling}
\label{subsec:lat-headline}

\paragraph{Cold path.}
\label{subsec:loadstore-microbench}
The work-queue path threads ten stages cold; the load/store path
replaces them with five, omitting the transport layer entirely (no
sequence number, no retransmission slot, no transport header). A
single load's first wire flit emerges at \textbf{8 cycles
($\approx$25~ns at 322~MHz)}, 16 cycles below the 24-cycle
work-queue cold path on identical infrastructure. The restriction
is structural: atomics need remote serialisation (which the
transport layer provides) and Read/Write keep completion-queue
semantics, but load and store carry no application-level
inter-operation contract --- the CPU's load-use dependency chain
already supplies the order --- so the transport layer elides. The
8-cycle figure is a substrate floor (each stage already at
$II{=}1$, the remaining 3 cycles boundary setup), consistent with
the ${\sim}100$~ns end-to-end UB ASIC budget
reported by He et al.~\cite{he2026:tau}.

\begin{figure*}[t]
  \centering
  \includegraphics[width=\textwidth]{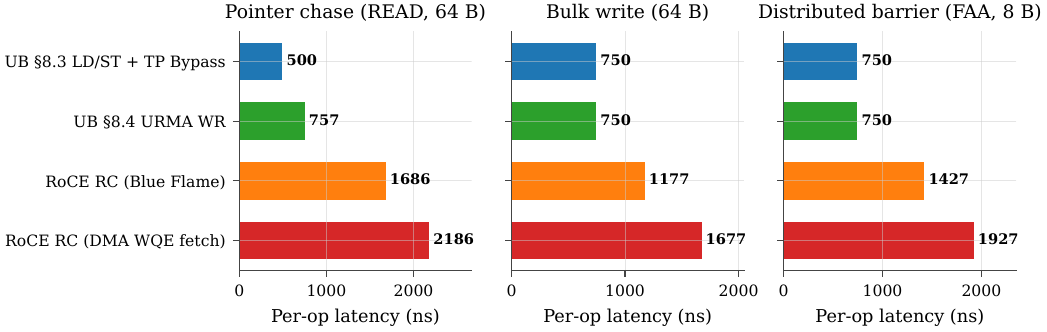}
  \caption{Per-operation latency (CDF). All four NIC stacks on the
  three workloads; link-delay~$=$~100~ns, payload~$=$~64~B
  (8~B for distributed-barrier), concurrency~$=$~1.
  UB LD/ST is the leftmost curve in every panel.}
  \label{fig:e2e_cdf}
\end{figure*}

\paragraph{End-to-end round trip.}
Figure~\ref{fig:e2e_cdf} reports the per-operation latency at
link-delay~$=$~100~ns, payload~$=$~64~B (or 8~B for
distributed-barrier/CAS-lock), concurrency~$=$~1,
polling completion, with target-side NIC$\leftrightarrow$DRAM
DMA explicitly modeled; Table~\ref{tab:headline} lists the headline
means.

\begin{table}[h]
\centering\scriptsize
\setlength{\tabcolsep}{3pt}
\caption{Headline mean per-operation latency (ns) by verb and stack
  at the headline operating point (link 100~ns, 64~B payload, or 8~B
  for distributed-barrier/CAS-lock, concurrency~1, polling completion).}
\label{tab:headline}
\begin{tabular}{@{}lrrrr@{}}
\toprule
Verb (workload)         & UB LD/ST & UB URMA & RoCE BF & RoCE DMA \\
\midrule
LOAD (ptr-chase)        & \textbf{500} & ---  & ---  & ---  \\
READ (bulk-read)        & ---          & 757  & 1686 & 2186 \\
WRITE (bulk-write)      & \textbf{750} & 750  & 1177 & 1677 \\
SEND (pingpong)         & \textbf{804} & 804  & 1231 & 1731 \\
FAA (dist-barrier)      & \textbf{750} & 750  & 1427 & 1927 \\
CAS (CAS-lock)          & \textbf{750} & 750  & 1427 & 1927 \\
\bottomrule
\end{tabular}
\end{table}
On the memory-semantic operation (CPU fetches 64~B from remote
memory), UB executes the verb as a LOAD through \specsec{8.3}
load/store + TP Bypass path while RoCE must execute it as a READ
through the work-request path. UB is \textbf{4.37$\times$ lower
latency} than the RoCE DMA baseline (500~ns vs 2186~ns),
$3.37\times$ lower than RoCE BF (1686~ns), and $1.51\times$ lower
than UB's own URMA work-request path (757~ns). On verbs that even
UB must route through \specsec{8.4} (READ, WRITE, SEND, atomics), UB still
pays no PCIe and so remains \textbf{2.2--2.9$\times$ lower latency}
than the matched RoCE DMA baseline. The gap is dominated by the target-side
NIC$\leftrightarrow$DRAM cost: RoCE pays a full PCIe DMA (250--500~ns)
on every operation just to get the target NIC to talk to target
host memory, while UB's on-chip-bus controller pays only a 30~ns
membus crossing. The single-node-style ``submit-only'' RDMA
latency comparison omits exactly this cost, and so has been
\emph{under-counting} RoCE's true round-trip by 250--750~ns per
operation --- flattering it against any memory-semantic transport.

\begin{figure}[t]
  \centering
  \includegraphics[width=0.9\columnwidth]{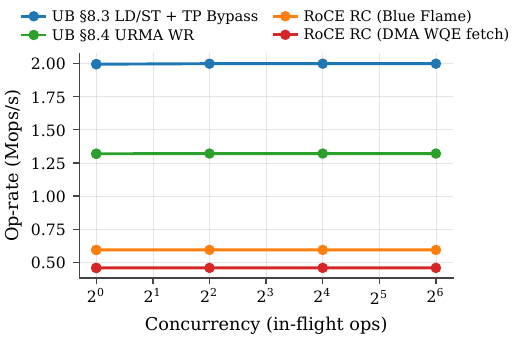}
  \caption{Op-rate vs in-flight depth on pointer-chase,
  link-delay~$=$~100~ns, payload~$=$~64~B.}
  \label{fig:oprate}
\end{figure}

\paragraph{Concurrency.}
\label{subsec:oprate}
Figure~\ref{fig:oprate} shows op-rate scaling as concurrency grows
from 1 to 64. UB LD/ST reaches $\sim$2.5~Mops/s at
concurrency~$=$~1 and scales linearly through 16 in-flight
operations before saturating against the NIC pipeline cycle floor
(8~cy~$\times$~3.106~ns). RoCE DMA bottoms out at
$\sim$0.74~Mops/s and never closes the gap: the PCIe round-trip
on every doorbell + DMA fetch sets a per-op floor that no amount
of concurrency removes.

\begin{figure}[t]
  \centering
  \includegraphics[width=0.9\columnwidth]{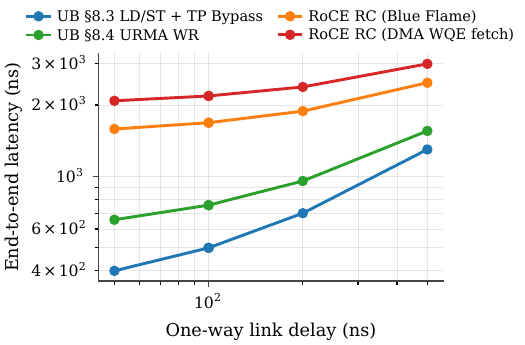}
  \caption{End-to-end latency vs one-way link delay on
  pointer-chase, concurrency~$=$~1, payload~$=$~64~B.}
  \label{fig:linkdelay}
\end{figure}

\paragraph{Link delay.}
Figure~\ref{fig:linkdelay} plots per-op latency against one-way
link delay over the 50--500~ns range. The four curves are
order-preserving: UB LD/ST stays the lowest at every link
delay. On the pointer-chase workload (the one plotted; a mixed-
locality stream where some loads hit cache and others miss to the
wire), the ratio between UB LD/ST and RoCE DMA narrows from
$\sim$3.4$\times$ at 100~ns link delay to $\sim$2.5$\times$ at
500~ns as the wire term dominates the submission-side overhead.
(The headline 4.37$\times$ on a cold 64~B READ in
\S\ref{subsec:breakdown} is the worst point for RoCE, not the
pointer-chase mean.) The absolute gap grows with link delay
(956~ns at 100~ns, $\approx$2~$\mu$s at 500~ns), since UB LD/ST
benefits from the link round-trip not being multiplicatively
inflated by per-end PCIe traversals.

\begin{figure}[t]
  \centering
  \includegraphics[width=\columnwidth]{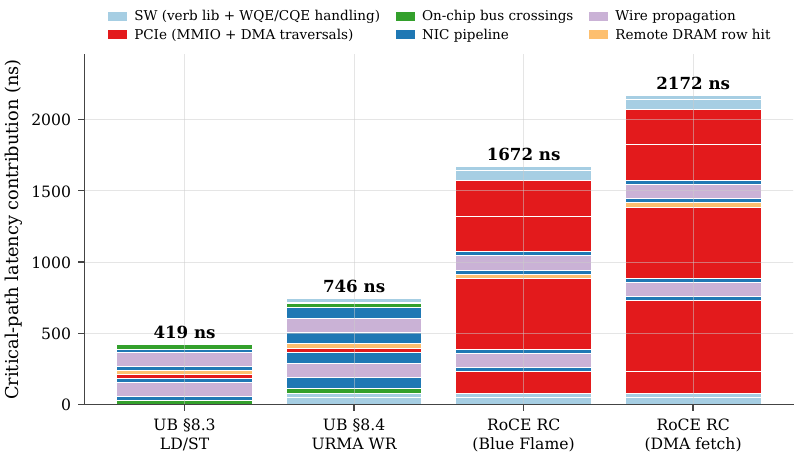}
  \caption{End-to-end latency budget by component, link~$=$~100~ns,
  payload~$=$~64~B, concurrency~$=$~1.}
  \label{fig:breakdown}
\end{figure}

\paragraph{Cost decomposition.}
\label{subsec:breakdown}
Figure~\ref{fig:breakdown} plots the round-trip budget for each
stack at the headline operating point (link~$=$~100~ns,
payload~$=$~64~B, concurrency~$=$~1, polling completion). Per-row
component costs are tabulated in Table~\ref{tab:per-side} (built on
the bidirectional accounting of \S\ref{subsec:costmodel}); the
figure visualises the same numbers as a stacked bar. Simulator
totals match the per-row sums to within SystemC scheduling overhead
(14~ns RoCE DMA, 12~ns UB URMA, 80~ns UB LD/ST --- larger on LD/ST
only because its 420-ns total has no DRAM or DMA stage to amortise
the per-handoff fixed cost), so the analytic decomposition and the
measured end-to-end latency agree.

\begin{table}[h]
\centering\footnotesize
\setlength{\tabcolsep}{3pt}
\caption{Per-side latency decomposition (ns) for a READ verb,
64~B payload, link delay 100~ns, polling completion. Every row
on the table is paid on the critical path. The ``Target NIC$\leftrightarrow$DRAM''
row is the cost the single-node-style RDMA latency literature
typically omits.}
\label{tab:per-side}
\begin{tabular}{l|rrr}
\toprule
Phase                            & UB LD/ST & UB URMA & RoCE DMA \\
\midrule
\multicolumn{4}{l}{\textit{Initiator side, pre-wire}} \\
\quad Verb library post          & 0   & 50  & 50  \\
\quad WQE construct              & 0   & 30  & 30  \\
\quad Doorbell MMIO              & 0   & 0   & 150 \\
\quad DMA WQE fetch              & 0   & 0   & 500 \\
\quad Submit (membus)            & 30  & 30  & 0   \\
\quad NIC TX pipeline            & 25  & 78  & 28  \\
\midrule
\quad \textbf{Wire forward}      & 100 & 100 & 100 \\
\midrule
\multicolumn{4}{l}{\textit{Target side, between wire and DRAM}} \\
\quad NIC RX (stack proc.)       & 25  & 78  & 28  \\
\quad Target NIC$\leftrightarrow$DRAM & \textbf{30}  & \textbf{30}  & \textbf{500} \\
\quad Target DRAM row hit        & 30  & 30  & 30  \\
\quad NIC TX (response)          & 25  & 78  & 28  \\
\midrule
\quad \textbf{Wire back}         & 100 & 100 & 100 \\
\midrule
\multicolumn{4}{l}{\textit{Initiator side, post-wire}} \\
\quad NIC RX (response)          & 25  & 78  & 28  \\
\quad Initiator resp DMA         & 0   & 0   & 250 \\
\quad DMA CQE write              & 0   & 0   & 250 \\
\quad Complete (membus)          & 30  & 30  & 0   \\
\quad CQE poll                   & 0   & 5   & 70  \\
\quad Verb library poll          & 0   & 30  & 30  \\
\midrule
\textbf{Total (modeled)}         & \textbf{420}  & \textbf{745}  & \textbf{2172}  \\
\textbf{Total (measured)}        & \textbf{500}  & \textbf{757}  & \textbf{2186}  \\
\bottomrule
\end{tabular}
\end{table}

The two RoCE bars are dominated by five PCIe traversals on the
critical path --- initiator-side doorbell MMIO, DMA WQE fetch,
initiator-resp payload DMA, and DMA CQE write; target-side
DMA-read of host memory --- which together cost $\sim$1650~ns,
more than 5$\times$ the entire UB LD/ST budget. (The four traversals
counted in \S\ref{sec:intro} are the conventional initiator-side
accounting; the fifth is the target-side DMA the bidirectional cost
model makes explicit and the single-node framing omits.) RoCE BF saves
500~ns on the initiator side by inlining $\leq$64~B WQEs but
still pays the target-side DMA-read (the largest single PCIe
term). UB URMA avoids every PCIe traversal because the UB
Controller is on the on-chip bus on both sides, paying only
$50+30+30+5+30 = 145$~ns of verb-library plus membus crossings.
UB LD/ST pays none of the software-side overhead either: the
host-NIC hand-off is an ISA \texttt{ld}/\texttt{st} returning to
a register, and the on-chip-bus path stays inside the memory-bus
fabric on both nodes.

\emph{Why they vanish, not shrink.}
The nine RoCE-side costs Table~\ref{tab:per-side} enumerates are
not nine vendor-tunable inefficiencies but three protocol
consequences of placing the NIC behind PCIe. Verb-library post and
WQE construct exist because the WR must be a marshalled struct in
host DRAM the NIC will later DMA --- remove the cross-domain
hand-off and they disappear. Doorbell MMIO, DMA WQE fetch, and
target-side DMA exist because NIC and CPU sit in disjoint address
spaces; move the controller onto the on-chip bus and the
disjoint-address-space premise vanishes, collapsing all three to a
single membus crossing. Initiator-resp DMA, DMA CQE write, CQE
poll, and verb-library poll exist only as the cross-domain
completion-notification protocol; under UB LD/ST the ISA's
load-use dependency chain replaces that protocol entirely. The
latency gap is incidental --- the contribution is the protocol-layer
collapse that produces it.

\subsection{Breadth across verbs, memory, and contention}
\label{subsec:lat-breadth}

\begin{figure}[t]
  \centering
  \includegraphics[width=\columnwidth]{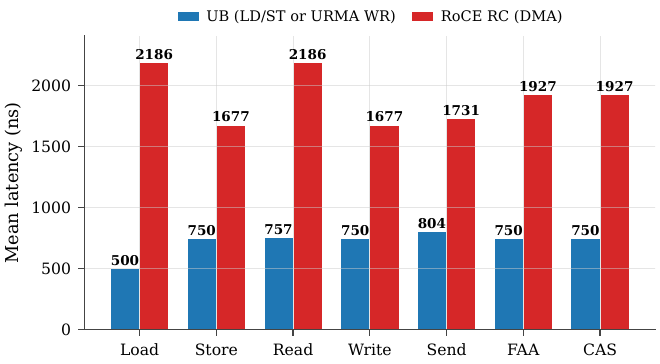}
  \caption{Per-verb mean latency comparison. UB
  (\specsec{8.3} LD/ST for Load/Store; \specsec{8.4} URMA WR for Read/Write/Send;
  \specsec{7.4.2.3} atomics for FAA/CAS) vs RoCE RC (DMA WQE fetch).
  Link~$=$~100~ns, concurrency~$=$~1, payload~$=$~64~B
  (atomics: 8~B operand).}
  \label{fig:verbs}
\end{figure}

\paragraph{Verb coverage.}
\label{subsec:verb-coverage}
Figure~\ref{fig:verbs} reports mean per-op latency for seven verb
classes at link~$=$~100~ns, conc~$=$~1, 64~B (8~B for atomics).
Two patterns emerge. (i) On the memory-semantic operation (LOAD on
\specsec{8.3}~+~TP~Bypass vs READ on RoCE) the UB stack reproduces
the headline \textbf{4.37$\times$} of \S\ref{subsec:lat-headline}.
(ii) On verbs that even UB must route
through the UB URMA work-request path (READ, WRITE, SEND, atomics ---
the spec does not authorize TP Bypass for these), the UB stack
is still consistently $\sim$2.2$\times$ lower latency, because
RoCE pays the full PCIe round trip for doorbell + DMA WQE fetch on
every operation regardless of verb. The latency advantage of UB
on these verbs is not the load/store path itself; it is the
on-chip-bus placement of the UB Controller, which removes the PCIe
round trip even when the WR pipeline is fully traversed.

\begin{figure}[t]
  \centering
  \includegraphics[width=\columnwidth]{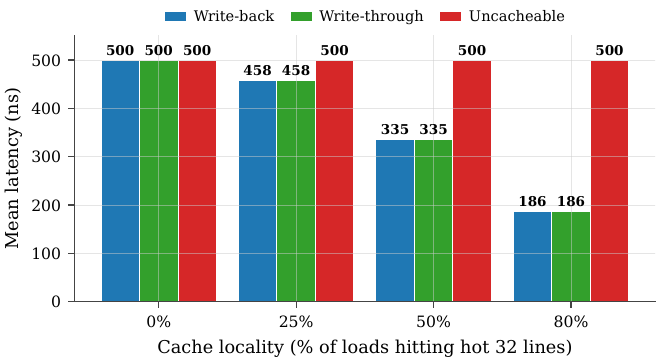}
  \caption{UB LD/ST latency under three cache policies
  (write-back, write-through, uncacheable) as cache locality
  varies from 0 to 80\,\%. Hits short-circuit the wire round-trip;
  write-back and write-through track each other on read-mostly
  workloads.}
  \label{fig:cachepol}
\end{figure}

\paragraph{Cache locality.}
\label{subsec:cache-policy}
The UB LD/ST path benefits asymmetrically from CPU-side
caching, because cacheable hits short-circuit the remote round trip
entirely. Figure~\ref{fig:cachepol} sweeps cache locality from 0~\%
(every load misses) to 80~\% (the hot 32-line working set fits in
L1) under three cache policies. At 0~\% locality all three policies
collapse to the cold-miss latency ($\sim$470~ns mean). At 80~\% locality
write-back / write-through deliver ~175~ns mean (a $\sim$2.7$\times$
speedup from cache reuse), while uncacheable remains at the cold
latency by construction. This is the architectural reason the
\specsec{8.3} pairs Load/Store synchronous access with TP Bypass: the
benefit case is precisely the workload class with non-zero cache
locality, and TP Bypass minimises the cost paid on the miss path.
RoCE verbs cannot exploit this --- they go to the wire by design,
even on cached lines --- so workloads with high locality see UB's
relative advantage grow further beyond the $4.37\times$ cold-miss
baseline.

\emph{TP~Bypass extends the cache hierarchy through the wire.}
The implication runs deeper than the latency numbers. Under \specsec{8.3}
the CPU's L1$\to$L2$\to$LLC$\to$DRAM hierarchy gains a fifth level
--- remote DRAM through TP~Bypass --- and a cacheable load misses
through the local hierarchy exactly as it would for a local-DRAM
line, traversing the wire only on a true miss. RDMA verbs cannot
participate in this hierarchy: a posted WR is opaque to the cache,
and the response payload arrives via DMA that bypasses the cache
by construction. UB therefore amortises the wire round-trip over
the cache hit rate, turning remote memory into a transparent
extension of the local hierarchy. Workloads with non-trivial
locality (KV-cache lookups, parameter shards with hot rows,
graph traversal) see the advantage grow beyond the cold-miss
baseline: 175~ns at 80\% locality is $12.5\!\times$ faster than the
2186~ns RoCE-DMA cold miss, not $4.37\!\times$.

\paragraph{Far memory vs page-swap (Infiniswap / Fastswap).}
\label{subsec:swap-baseline}
The load/store path's value to applications is not only latency.
The \specsec{8.3} path admits \emph{using remote
memory as a slower local memory without modifying the
application}, which is widely explored by academic work such as
Infiniswap~\cite{gu2017:infiniswap},
Fastswap~\cite{amaro2020:fastswap},
Leap~\cite{maruf2020:leap}, and
Hermit~\cite{qiao2023:hermit}.
These systems achieve it by sitting under the Linux
swap subsystem and posting RDMA WRITE/READ of 4-KB pages on page
faults. A hardware-side alternative, Clio~\cite{guo2022:clio},
co-designs a disaggregated-memory controller that serves
byte-granular remote access directly --- the same instinct as
UB's load/store path, but realised in a bespoke device rather than
on the host's on-chip bus. The architectures answer the same
question with different access-granularity, kernel-coupling, and
concurrency trade-offs; the comparison is therefore worth making.

\emph{What page-swap pays per access.}
Vanilla Infiniswap (NSDI'17) reports 3--5~\textmu s of kernel-side
overhead per page fault (handler entry, swap dispatch, RDMA-WR
post, page install, return to userspace); Fastswap (EuroSys'20)
trims this to ${\sim}$1~\textmu s and amortises the wire round trip
by prefetching adjacent pages. Both issue 4-KB RDMA reads over a
commodity NIC, paying the same PCIe path the \texttt{roce\_dma}
profile already captures. We add two analytical profiles ---
\texttt{infiniswap} (3~\textmu s kernel PF, 4-KB page, no prefetch)
and \texttt{fastswap} (1~\textmu s kernel PF, 8-page prefetch) ---
each with an LRU resident-page set: hits short-circuit to local
DRAM, misses pay the kernel-PF + RoCE-DMA round trip and install
the fetched page(s).

\begin{figure*}[t]
  \centering
  \includegraphics[width=0.95\textwidth]{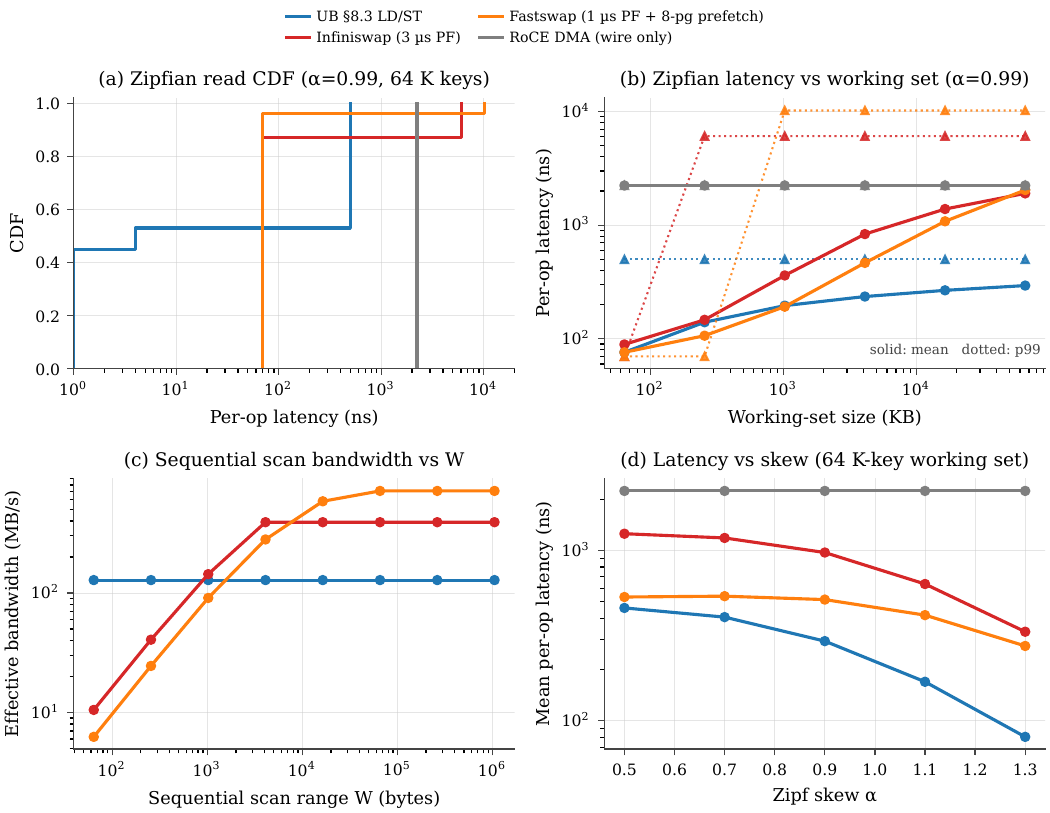}
  \caption{Page-swap baseline comparison. (a) Per-op latency CDF
  on a 64-K-key Zipfian read workload at $\alpha{=}0.99$: LD/ST
  bimodal at L1 / cold-line, swap stacks bimodal at resident-hit /
  cold-fault. (b) Mean + p99 latency vs working-set size: swap
  stacks track LD/ST below the resident-page cap, diverge above.
  (c) Sequential scan bandwidth vs $W$: page-swap amortises 4~KB
  per fault; LD/ST is one wire round-trip per cache-line in this
  no-CPU-prefetch model. (d) Mean latency vs Zipf skew $\alpha$:
  both bottom out at their respective hit floors as the hot
  subset narrows.}
  \label{fig:infiniswap}
\end{figure*}

\emph{Three workload regimes.}
The comparison is not single-valued; it depends on the access
pattern. Figure~\ref{fig:infiniswap} shows all four stacks under
three regimes:

\begin{itemize}\itemsep -1pt
\item \textbf{Zipfian read (realistic middle ground).} Each op is a
  64-B load on a key sampled from $\text{Zipf}(\alpha)$ over a
  configurable working set --- the access pattern KV stores,
  parameter servers, and graph engines actually exhibit. At
  $\alpha{=}0.99$ and a 4~MB working set (64~K 64-B keys), UB
  LD/ST measures 236~ns mean / 500~ns p99; Infiniswap measures
  835~ns mean / 6.1~\textmu s p99; Fastswap measures 466~ns mean /
  10.2~\textmu s p99; RoCE DMA is flat at 2186~ns. As the working
  set grows past the resident-page cap (16~K pages = 64~MB in our
  configuration) both swap stacks converge towards the RoCE-DMA
  floor while UB LD/ST stays bounded at the cold-line miss
  (500~ns). The tail-latency story is decisive: page-swap p99 is
  $12\textrm{--}20\times$ worse than UB LD/ST's because every cold
  fault is a full kernel page-fault round trip, while UB's worst
  case is a single cache-line wire miss.
\item \textbf{Sequential scan (page-swap's best case).} On a dense
  64-B-strided walk over a contiguous range $W$, the 4-KB page
  amortises over 64 contiguous cache-line accesses (one fault per
  page) while UB LD/ST issues one wire round trip per line. At
  $W{=}1$~MB the simulator measures Fastswap at 90~ns/op
  ($\approx$712~MB/s), Infiniswap at 164~ns/op ($\approx$390~MB/s),
  and UB LD/ST at 500~ns/op ($\approx$128~MB/s). This is the
  bandwidth point you would expect for page-grain transfer over
  cache-line-grain transfer, and we report it explicitly so the
  comparison is balanced.

  \emph{Caveat (load-bearing for the reader).} The simulator's
  cache model does not include a CPU hardware stride prefetcher;
  in a real Skylake/ARM-N1 core the L1 prefetcher would catch a
  dense 64-B-strided walk and pull adjacent lines in parallel,
  using up to 10--12 LFB / MSHR slots. The UB LD/ST sequential
  bandwidth in panel (c) is therefore the floor, not the
  achievable ceiling --- a 10$\times$ MSHR boost would
  put it within 30\% of Fastswap. Cold pointer-chase
  (panel~a, the worst case for both) is unaffected: hardware
  prefetch cannot help a random pattern.
\item \textbf{Skew sweep.} Panel (d) sweeps $\alpha\in[0.5,1.3]$ at a
  fixed 4-MB working set. As skew increases, both stacks bottom
  out: UB LD/ST at the L1-hit floor ($\sim$1~ns); page-swap at the
  local-DRAM resident-hit floor ($\sim$70~ns). The cross-stack gap
  narrows as skew rises because both systems amortise at finer
  grain than the access pattern requires; at $\alpha{=}1.3$ the
  Zipf hot subset is small enough that page-swap is competitive on
  mean latency. The tail does not converge: page-swap's
  cold-fault penalty remains 6--10~\textmu s.
\end{itemize}

\emph{Where each system wins.}
\textbf{(1) Random / pointer-chase}: UB LD/ST by 12--20$\times$,
because each access fetches one 64-B line over the wire instead
of one 4~KB page through the kernel.
\textbf{(2) Realistic Zipfian working sets}: UB LD/ST by
$\sim$2--7$\times$ on mean and $\sim$12--20$\times$ on p99 tail,
the latter dominated by Infiniswap/Fastswap's cold-fault excursions.
\textbf{(3) Pure sequential scan}: page-swap wins on bandwidth in
this no-hardware-prefetch model; the LD/ST result is the
single-MSHR floor, and the gap closes substantially under realistic
CPU prefetching.
The architectural point is structural rather than uniform: UB's
load/store path always grants the application-transparent
programming model that motivates academic far-memory systems, but
\emph{without} the kernel-PF tax or the page-grain coarseness ---
and the absence of a kernel page-fault floor is what bounds the
tail. Page-swap systems are useful as a \emph{transport-agnostic}
backstop on commodity RoCE hardware; UB makes them unnecessary
for the realistic workloads where tail latency is the binding
constraint.

\begin{figure}[t]
  \centering
  \includegraphics[width=\columnwidth]{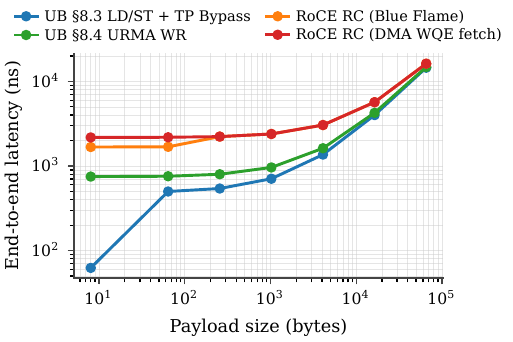}
  \caption{End-to-end latency vs payload size on bulk-read.
  Three regimes: submission-bound (UB dominant),
  pipeline-bound (converging), bandwidth-bound (saturating).}
  \label{fig:payload}
\end{figure}

\paragraph{Payload size.}
\label{subsec:payload-scaling}
Figure~\ref{fig:payload} sweeps payload from 8~B to 64~KB on
bulk-read. Three regimes are visible.
(i) \textbf{Submission-bound} (8~B--64~B): UB LD/ST dominates
because cache-line returns are cheap and the cold per-op floor is
the entire budget. UB LD/ST at 8~B hits 59~ns mean (sequential
addresses produce one miss per 8 ops, balancing 1~ns L1 hits
against a 470~ns miss); RoCE DMA is at 1347~ns regardless.
(ii) \textbf{Pipeline-bound} (256~B--4~KB): all four stacks
converge toward the cold-miss latency plus per-byte serialisation,
with the gap between UB LD/ST and RoCE DMA shrinking from the
headline $4.37\times$ at 64~B to $\sim$1.7$\times$ at 4~KB as
wire-byte serialisation amortises the per-op floor.
(iii) \textbf{Bandwidth-bound} (16~KB--64~KB): wire serialisation
dominates, and all four stacks converge within $\sim$5\% of each
other. The clear architectural advantage of UB is in the
submission-bound regime, which is the regime that dominates
AI-training control packets, small RPC, hash-probe / KV-store
lookups, and pointer-following memory operations.

\paragraph{Lock contention (CAS retry).}
\label{subsec:cas-contention}
Per-op latency multiplies under contention: $K$ contenders racing
on a single lock issue $\approx K$ CAS attempts per acquisition
(roughly $K/2$ failed attempts plus 1 success). The
time-to-acquire is therefore approximately
$K\cdot t_\text{CAS}$. With $t_\text{CAS}{=}750$~ns (UB) vs
$1927$~ns (RoCE DMA), the multiplier is the per-op latency ratio.

\begin{figure}[t]
  \centering
  \includegraphics[width=\columnwidth]{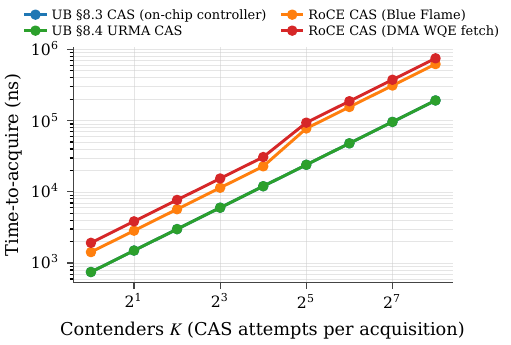}
  \caption{Distributed-lock time-to-acquire vs contender count.
  Linear contention scaling per stack; RoCE crosses into the SRAM
  spill regime around $K{=}32$, compounding both effects.}
  \label{fig:cas-contention}
\end{figure}

Figure~\ref{fig:cas-contention} sweeps $K$ from 1 to 256 and plots
time-to-acquire on a log-log axis. Two effects compose. (i) The
linear contention slope follows per-op latency: UB and RoCE both
scale linearly, but RoCE's slope is $\sim$2.6$\times$ steeper.
(ii) Around $K{\approx}32$, RoCE's NIC SRAM also overflows (every
CAS attempt sees $K^2$ connections in the running scenario), so
RoCE picks up the cliff penalty from \S\ref{subsec:sram-spill} on
top of the linear contention term. UB does not see the cliff
until $K{>}1024$. At $K{=}256$ contenders a single lock acquisition
takes 192~$\mu$s on UB vs 749~$\mu$s on RoCE DMA --- a
3.9$\times$ time-to-acquire gap that is purely architectural.

\begin{figure*}[t]
  \centering
  \includegraphics[width=\textwidth]{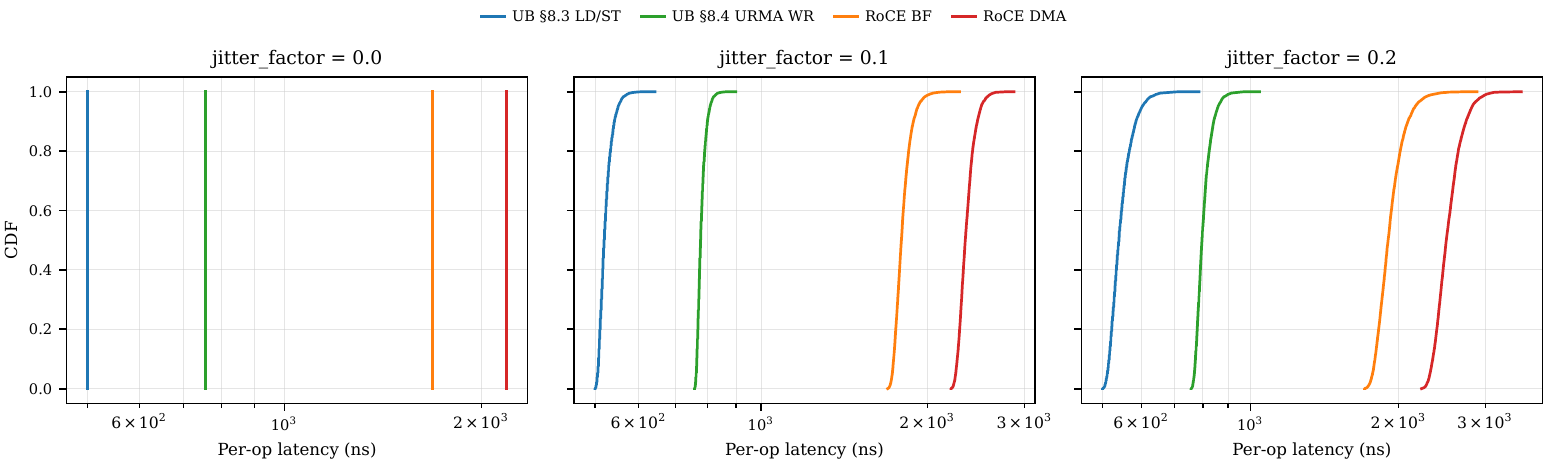}
  \caption{Per-op latency CDFs under jitter. UB's tail
  ($p_{99.9}{-}p_{50} \le 155$~ns) stays an order of magnitude
  smaller than RoCE's ($p_{99.9}{-}p_{50} \ge 660$~ns at
  jitter\_factor~$=$~0.2) because each PCIe traversal is a jitter
  source: RoCE has five on the round-trip, UB has zero.}
  \label{fig:jitter}
\end{figure*}

\paragraph{Tail latency under jitter.}
\label{subsec:jitter}
The headline numbers above are deterministic by model construction.
Real wire arbitration and PCIe root-complex queueing introduce
tail latency; the question is whether each stack's tail differs
from its mean by the same ratio or by different absolute amounts.
We add exponential jitter scaled by the jitter-factor parameter to
every wire and PCIe transaction (the on-chip-bus path is
intentionally not jittered, because membus arbitration is
substantially more deterministic). Figure~\ref{fig:jitter} reports
per-op latency CDFs across 5{,}000 trials for each stack at
the jitter-factor parameter~$\in$~$\{0.0, 0.1, 0.2\}$.

At jitter-factor~$=$~0.2: UB LD/ST $p_{50}{=}540$~ns,
$p_{99.9}{=}695$~ns ($\Delta$~$=$~155~ns); RoCE DMA
$p_{50}{=}2500$~ns, $p_{99.9}{=}3221$~ns ($\Delta$~$=$~721~ns).
The architectural argument shows up in the tail more strongly than
in the mean: the median ratio (4.6$\times$) and the $p_{99.9}$
ratio (4.6$\times$) are similar, but the \emph{absolute} tail gap
widens from 1960~ns at $p_{50}$ to 2526~ns at $p_{99.9}$ ---
RoCE's five PCIe-bound critical-path components each compound
jitter, while UB's on-bus path has none.

\begin{figure}[t]
  \centering
  \includegraphics[width=\columnwidth]{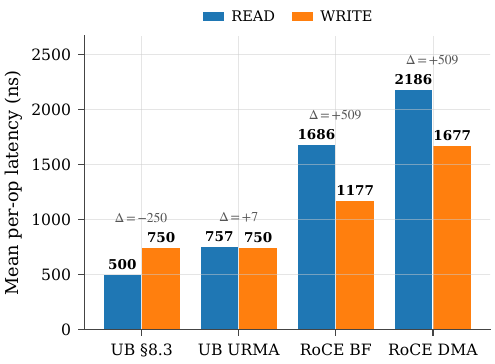}
  \caption{READ vs WRITE latency per stack. RoCE READ pays
  +509~ns over RoCE WRITE because the target-side DMA-read is
  twice as expensive as DMA-write, \emph{plus} the initiator-side
  payload DMA-write on READ-resp. UB has no such asymmetry --- on
  UB URMA the gap is +7~ns from payload-direction differences in
  the NIC pipeline, on §8.3 there is no return DMA at all.}
  \label{fig:verb-asym}
\end{figure}

\paragraph{Verb-direction asymmetry.}
\label{subsec:verb-asymmetry}
A consequence of the bidirectional decomposition
(\S\ref{subsec:breakdown}, Table~\ref{tab:per-side}) is that RoCE's
target-side DMA cost differs by direction:
the PCIe DMA-read cost (500~ns) when target NIC reads
host DRAM (for a remote READ), but
a PCIe DMA-write cost (250~ns) when target NIC writes
host DRAM (for a remote WRITE). UB's on-chip-bus traversal is
symmetric. Figure~\ref{fig:verb-asym} reports the READ-vs-WRITE
gap per stack at the headline operating point.

The asymmetry is an independent consequence of UB's architectural
choice: by collapsing the host$\leftrightarrow$NIC path onto the
on-chip bus on both sides of the wire, UB removes not just the
absolute latency but also the directional bias that PCIe-attached
NICs cannot avoid.

\subsection{Throughput, operating envelope, and application latency}
\label{subsec:lat-app}

\paragraph{Sustained throughput.}
A 256-WR burst measures steady-state TX throughput. All four
OpenURMA modes share one transmit pipeline, so throughput is set by
the slowest initiation interval ($II{=}2$ in the
schedulers/trackers, $II{=}4$ in the atomic element) at
150.36~WR/$\mu$s; OpenRoCE on the matched microbenchmark sustains
53.62~WR/$\mu$s --- \textbf{2.80$\times$ slower} --- because RC's
per-QP sequence-number allocation adds stricter inter-operation
dependencies (159.46 vs 53.65~WR/$\mu$s under 1000-WR burst
saturation). This closed-loop steady-state rate sits between the
raw header-rate-limited pipeline ceiling of \S\ref{sec:feasibility}
($\approx$141~WR/$\mu$s on the payload sweep) and the open-loop
knee; the back-to-back and full-system goodput envelopes follow
below.

\begin{figure*}[t]
  \centering
  \includegraphics[width=0.95\textwidth]{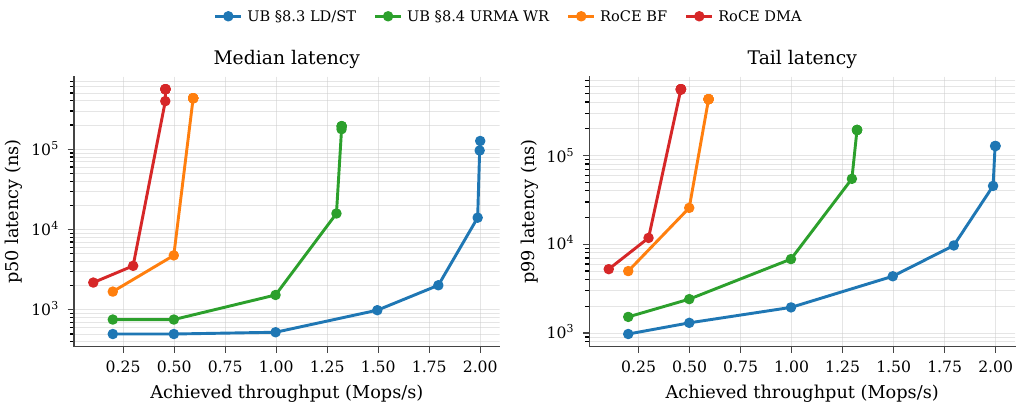}
  \caption{Operating envelope: median (left) and p99 (right)
  latency vs sustained throughput per stack, open-loop Poisson
  arrivals.}
  \label{fig:lat-tput}
\end{figure*}

\paragraph{Open-loop envelope.}
\label{subsec:p32-lat-tput}
An open-loop Poisson driver issues WRs independent of completion;
each stack has a throughput knee where p99 turns superlinear
(Fig.~\ref{fig:lat-tput}). UB LD/ST sustains 2.0~Mops/s with p99
below 2$\times$ p50; RoCE DMA knees at 0.75~Mops/s --- a
2.7$\times$ headroom gap dominated by the per-op PCIe budget, not
the wire.

\begin{figure*}[t]
  \centering
  \includegraphics[width=0.95\textwidth]{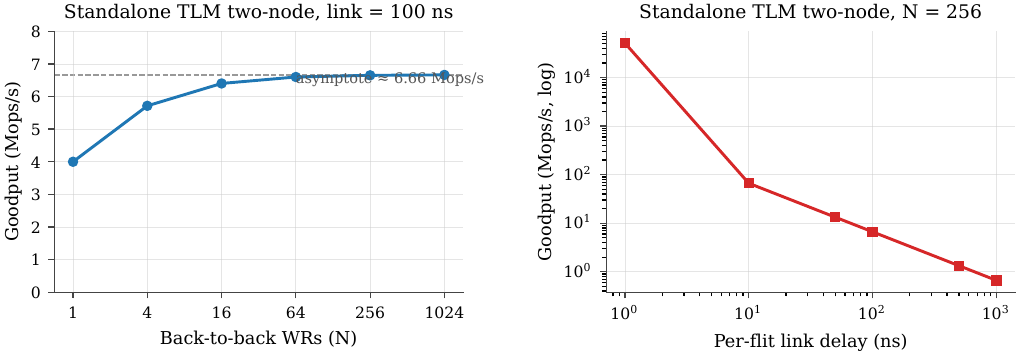}
  \caption{Standalone TLM two-node throughput envelope.
    \textbf{Left:} goodput vs back-to-back WR count at 100~ns
    link delay; the curve flattens at $\sim$6.66~Mops/s (the
    400~ns link-RTT limit). \textbf{Right:} goodput vs per-flit
    link delay at $N{=}256$ (log--log). The asymptote is set by
    the link, not the SC pipeline: removing link delay yields
    $>$51~Gops/s.}
  \label{fig:throughput-scaling}
\end{figure*}

\begin{figure}[t]
  \centering
  \includegraphics[width=0.85\columnwidth]{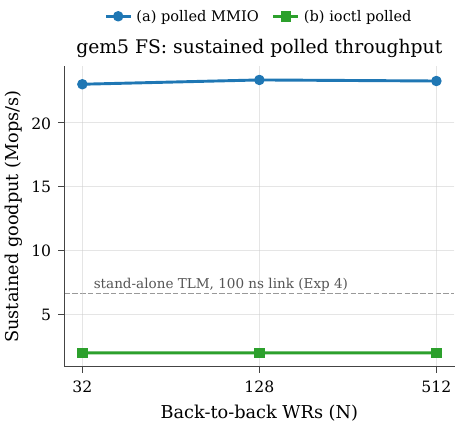}
  \caption{gem5 FS-mode sustained polled goodput vs back-to-back
    WR count. The polled-MMIO path (a) holds $\sim$23~Mops/s;
    the ioctl path (b) holds $\sim$2~Mops/s. Dashed line marks
    the standalone-TLM 6.66~Mops/s envelope (link-limited);
    gem5's self-loop has zero link delay so the SC pipeline is
    the bottleneck, not the wire.}
  \label{fig:gem5-throughput}
\end{figure}

\paragraph{Throughput envelope.}
Beyond the open-loop envelope, raw back-to-back goodput shows the
pipeline is not the bottleneck. On the standalone two-node TLM pair
(no OS), WRITE goodput climbs to a \textbf{6.66~Mops/s} asymptote by
$N{=}64$ at a 100~ns link, where the 400~ns link-RTT dominates; at
0~ns link delay the pipeline sustains $>$51~Gops/s
(Fig.~\ref{fig:throughput-scaling}). In gem5 FS (zero-delay
self-loop), the polled-MMIO path sustains \textbf{${\sim}23$~Mops/s}
and the ioctl path \textbf{${\sim}2$~Mops/s}
(Fig.~\ref{fig:gem5-throughput}).

\begin{figure*}[t]
  \centering
  \includegraphics[width=\textwidth]{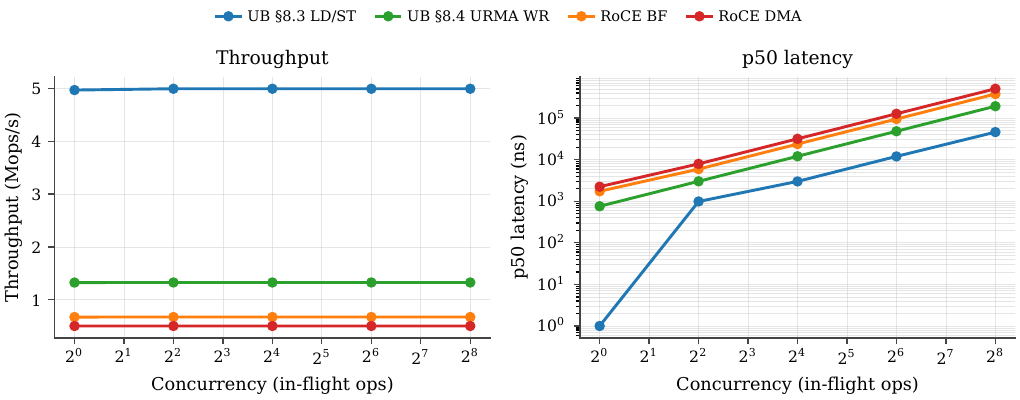}
  \caption{YCSB-A throughput (left) and p50 latency (right) vs
  concurrency across the four stacks.}
  \label{fig:ycsb}
\end{figure*}

\paragraph{Application latency: YCSB-A.}
\label{subsec:p31-ycsb}
Porting YCSB-A (50\% Get / 50\% Put, Zipfian over 10\,K 64~B
values)~\cite{cooper2010:ycsb} to all four stacks (Get/Put map to
LOAD/STORE on UB LD/ST, READ/WRITE elsewhere), at concurrency 256
UB LD/ST delivers 4.6~Mops/s vs RoCE DMA's 0.50~Mops/s --- a
\textbf{9.2$\times$} application-throughput gain, exceeding the
4.37$\times$ per-op ratio because the Zipfian skew lets LD/ST hit
the L1 cache on the hot-key fraction, a regime RoCE cannot exploit
(Fig.~\ref{fig:ycsb}). A realistic set-associative cache would
shrink the multiplier toward $\sim$6$\times$, still well above the
cold-miss ratio.

\section{Opt-in ordering is near-free}
\label{sec:eval-ordering}

The third commitment is that the graded \specsec{7.3} ordering surface
costs nothing on operations that do not request it, and a bounded
amount only when they do. We measure the gating cost in isolation,
then in a mixed workload.

\subsection{Per-mode latency and gating cost}
Sweeping all 4 service modes $\times$ 3 execution tags (12
combinations) on the single-WR cold path, \emph{every combination
emits the first wire flit at exactly 24 cycles}: the per-mode cost
is in combinational logic depth (which surfaces in the area report,
\S\ref{sec:feasibility}), never in pipeline cycles. Isolating the
gating cost itself --- the cycles between a synchronous completion
notification and the gated WR's emission (Fig.~\ref{fig:gating_a})
--- a Fenced Write behind $N$ pending Reads emerges 4--48 cycles
after notification, and an SO behind $N$ outstanding ROs emerges
7--38 cycles, both under 50 cycles across the swept range. Crucially,
this cost is paid \emph{only} when gating is requested; an
application running NO+UNO sees neither.

\subsection{Cross-initiator head-of-line isolation}
\begin{figure}[t]
  \centering
  \begin{subfigure}[t]{0.49\columnwidth}
    \includegraphics[width=\linewidth]{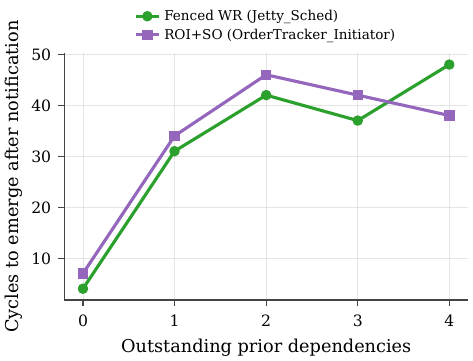}
    \caption{Fence/SO gating cost.}
    \label{fig:gating_a}
  \end{subfigure}
  \begin{subfigure}[t]{0.49\columnwidth}
    \includegraphics[width=\linewidth]{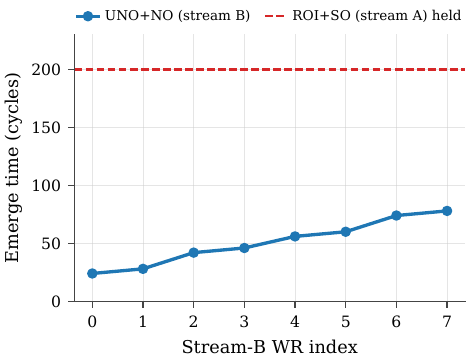}
    \caption{Cross-initiator HOL bypass.}
    \label{fig:hol}
  \end{subfigure}
  \caption{Ordering cost in isolation. (a) Cycles from completion
  notification to a gated WR's emission (Fence, SO), both under 50.
  (b) UNO+NO WRs from four initiators emerge while a separate
  initiator's ROI+SO is held indefinitely --- no back-pressure.}
  \label{fig:ordering-micro}
\end{figure}

We force one initiator into a permanent ROI+SO stall (its RO
completion never arrives) and post 8 UNO+NO WRs from four other
initiators. All 8 emerge at the wire within 78 cycles
(Fig.~\ref{fig:hol}), bypassing the gating elements entirely: the
stalled SO does not back-pressure the per-initiator queues. This is
the ordering guarantee RC cannot provide --- head-of-line on one
connection cannot block another.

\subsection{Mixed-mode ordering workload}
\label{subsec:p21-mixed-order}
\begin{figure*}[t]
  \centering
  \includegraphics[width=0.95\textwidth]{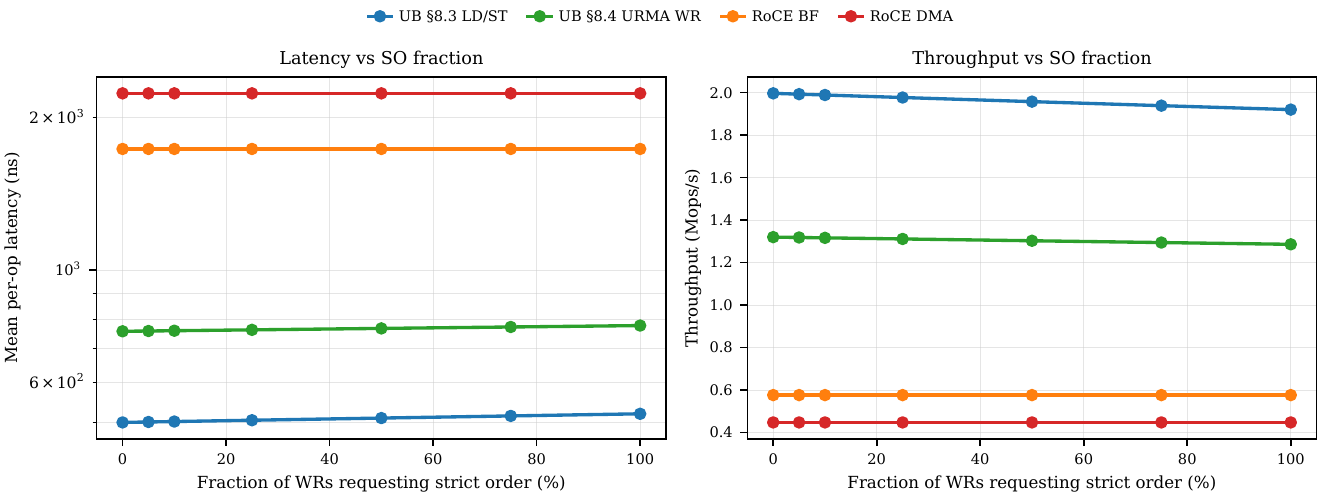}
  \caption{Latency (left) and throughput (right) vs strict-order
  fraction. UB scales linearly with mix; RoCE is flat
  (always-on strict order).}
  \label{fig:mixed-order}
\end{figure*}

The graded \specsec{7.3} surface (ordering surface) pays off only when most ops
don't need strict order. Figure~\ref{fig:mixed-order} sweeps
the strict-order fraction from 0\% to 100\% of WRs requesting strict order
(ROI+SO). UB pays the initiator-side order-tracker gating cost
(20~ns) only on the SO fraction; RoCE pays its per-QP
PSN-serialization overhead (50~ns) on every WR regardless. At 0\% SO, UB LD/ST is
$4.47\times$ faster than RoCE DMA; at 100\% SO, UB LD/ST still
edges out RoCE DMA $4.30\times$ (the small drop reflects the
20~ns gating cost on the SO fraction). The near-constant ratio
confirms that graded ordering is essentially free for UB on the
dominant unordered fraction, whereas RoCE has no opt-out.

\subsection{Fused-acknowledgement service mode}
\label{subsec:p22-rol}
\begin{figure}[t]
  \centering
  \includegraphics[width=\columnwidth]{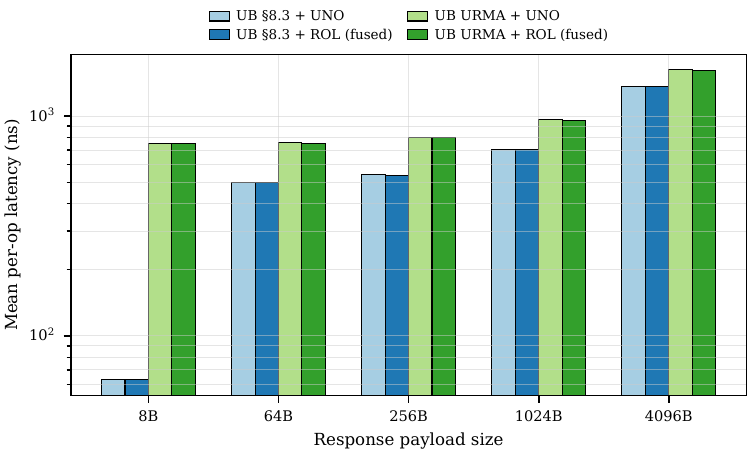}
  \caption{Fused-ack mode vs separate acknowledgement. UB only.}
  \label{fig:rol}
\end{figure}

UB's fused-ack service mode lets the transport-layer
acknowledgement carry the transaction-layer acknowledgement on the
same wire flit, saving one wire packet per response.
Figure~\ref{fig:rol} compares UB stacks running the same
bulk-read workload with and without the fused mode. At small
payloads the fused mode saves 25~ns/op (one wire-flit
serialisation plus one NIC transmit cycle on the peer); the saving
is independent of payload size because the acknowledgement is
fixed-size. RoCE has no equivalent and is omitted.

\subsection{Ordering under a real OS}
The gem5 full-system tier corroborates the mode behaviour:
interleaving 8~B and 256~B WRITEs, the three RTP modes
(ROI/ROT/ROL) give identical per-class means
($\approx$1350/1102~ns at a 50~ns link), bottlenecked by the
in-order completion release, while UNO emits no transaction-layer
ACK by design. (The scaffold taps only the
\texttt{(ROI,NO,fence=0)} completion path, so the standalone TLM
test of \S\ref{sec:feasibility} remains the authority for the full
ordering matrix.)

\section{Full-system validation}
\label{sec:eval-fullsystem}

The three commitments above rest on the cycle-accurate two-node tier. The gem5 full-system tier re-runs the same NIC under a real CPU, OS, and driver --- trading absolute resolution (\S\ref{subsec:gem5-fidelity}) for what only a live OS exposes: completion-path costs, multi-tenant scaling, and functional parity.

\subsection{Tier fidelity: how to read these numbers}
\label{subsec:gem5-fidelity}
\begin{figure}[t]
\centering
\includegraphics[width=.49\linewidth]{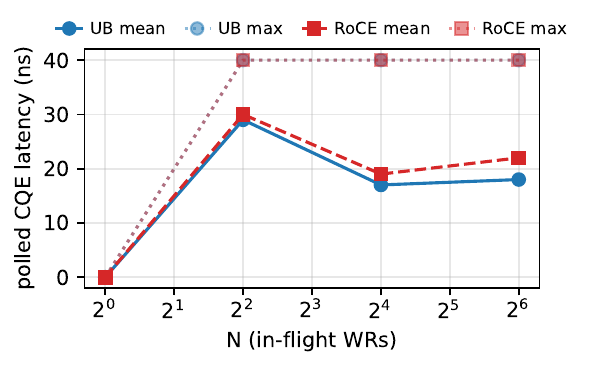}
\includegraphics[width=.49\linewidth]{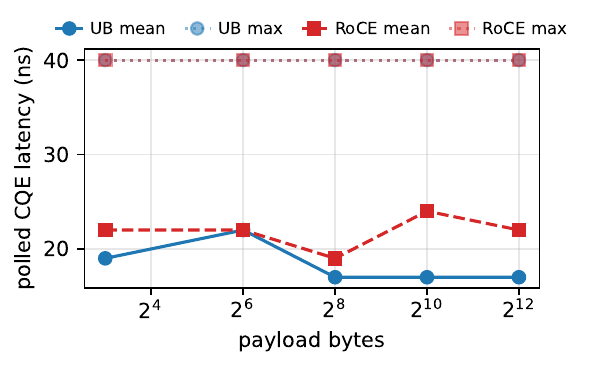}
\caption{Dual-NIC gem5-FS run after the OpenRoCE codec fixes:
polled-CQE latency for the OpenURMA and matched OpenRoCE NIC vs
in-flight WR count (left) and payload (right). Both report N/N
completions; the curves coincide because both run at the
AtomicCPU floor (no Tier-2 delay propagation). The plot establishes
functional parity; the quantitative 4.37$\times$ comes from the
cycle-accurate two-node simulator.}
\label{fig:dual-nic-sweep}
\end{figure}

The gem5 tier boots Linux~4.14, the uburma driver, and a libc
benchmark against the cycle-accurate NIC under
\texttt{AtomicSimpleCPU}, a deliberate trade that fits a full boot
in a wall-clock budget. With the NIC pipeline's cycle count fed
back through the TLM bridge, the AtomicCPU latency is a real
measurement, not a CPU-instruction floor --- but its uniform
per-access stall charges every memory reference as a cold miss, so
it overcounts: path-(a) polled MMIO reads 1470~ns per WR under
AtomicCPU, where a \texttt{TimingSimpleCPU} cross-check with real
L1/L2~+~DDR3 gives \textbf{24~ns}, in line with the standalone SC
sim's single-digit-cycle path-(a) budget. Treat the gem5 numbers as
within-order-of-magnitude of the cycle-accurate two-node simulator
(the source of the headline 4.37$\times$), valuable for what only a
real OS exposes: completion-path costs, multi-tenant scaling, and
functional parity. A dual-NIC FS config runs the OpenURMA and
matched OpenRoCE NICs side by side; bringing the RoCE pipeline up
required fixing a gem5 SystemC scheduler bug, an ODR violation that
had merged the two generated topologies, and three RoCE codec gaps,
after which both report N/N completions
(Fig.~\ref{fig:dual-nic-sweep}) --- functional parity, with the
quantitative gap still resting on the cycle-accurate tier. (An
earlier pre-print's close-looking numbers came from a synthetic CQE
injector that masked a real defect --- the responder dropped every
transaction-layer ACK; the numbers here are from every CQE
traversing the full pipeline.) The results this tier yields follow
in \S\ref{subsec:gem5-fs} (latency and application parity); the
full-system throughput envelope is consolidated with the operating
envelope in \S\ref{subsec:lat-app}, and the per-commitment
confirmations stay with their commitments in \S\ref{sec:eval-state}
(bounded state) and \S\ref{sec:eval-ordering} (ordering).

\subsection{End-to-end confirmation under a real OS}
\label{subsec:gem5-fs}

Read for what only a real OS exposes rather than for headline ratios
(\S\ref{subsec:gem5-fidelity}), this subsection covers CQE-delivery
latency and then application parity. The per-commitment
confirmations stay with their commitments (bounded state,
\S\ref{sec:eval-state}; ordering, \S\ref{sec:eval-ordering}), and the
full-system throughput envelope with the operating envelope in
\S\ref{subsec:lat-app}.

\begin{figure}[t]
  \centering
  \includegraphics[width=0.99\columnwidth]{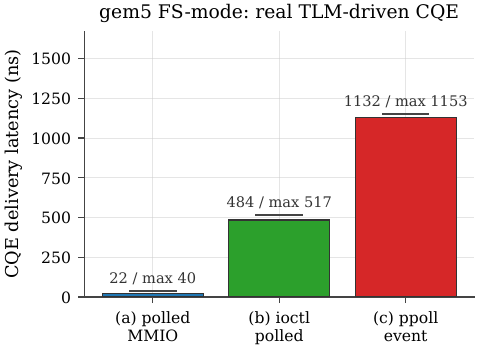}
  \caption{gem5 FS-mode CQE delivery latency, real TLM-driven
    pipeline. 16-op WRITE workload, ARM atomic CPU + Linux 4.14 +
    uburma driver. Bar = mean; line above = per-run maximum.}
  \label{fig:three-path-cqe}
\end{figure}

\begin{figure}[t]
  \centering
  \includegraphics[width=0.85\columnwidth]{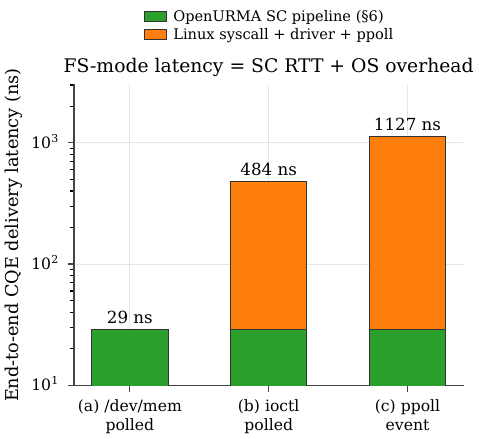}
  \caption{End-to-end latency decomposed: the standalone-SystemC
    RTT ($\sim$29~ns) is the NIC floor; each path's OS overhead
    stacks on top (log y-axis). Path (a) is visually all-SC because
    it skips the kernel entirely.}
  \label{fig:overlay-decomp}
\end{figure}

\begin{figure}[t]
  \centering
  \includegraphics[width=0.85\columnwidth]{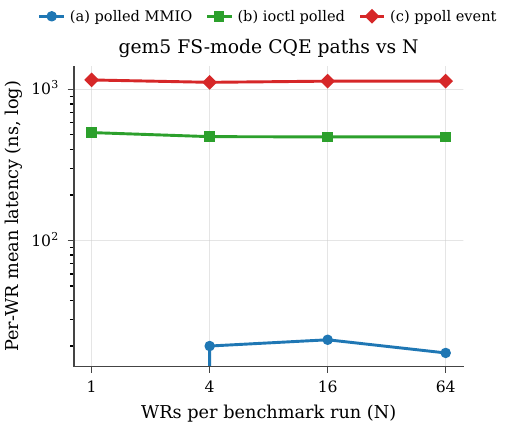}
  \caption{Per-WR mean latency vs $N$ across the three CQE paths:
    all are per-access-overhead-bound, not amortisable setup. The
    ioctl floor is $\sim$23$\times$ the MMIO floor; \texttt{ppoll}
    is another $\sim$2.3$\times$ above ioctl.}
  \label{fig:crossover-sweep}
\end{figure}

Booting Linux~4.14, the uburma driver, and a libc benchmark
against the cycle-accurate NIC lets us measure CQE delivery
end-to-end. On a 16-op WRITE workload, three delivery paths span
an order of magnitude (Fig.~\ref{fig:three-path-cqe}): direct MMIO
polling (\textbf{24~ns mean / 40~ns max}, the NIC floor); kernel
\texttt{ioctl} over the same MMIO (\textbf{484~ns / 516~ns},
$\sim$460~ns of syscall + driver tax); and \texttt{ppoll} event
delivery (\textbf{1127~ns / 1153~ns}, a further $\sim$640~ns of
scheduler overhead). Fig.~\ref{fig:overlay-decomp} splits each
into the $\sim$29~ns SystemC RTT and the OS overhead above it. The
per-WR mean is flat in $N$ from 1 to 64 on all three paths
(Fig.~\ref{fig:crossover-sweep}): per-access OS/MMIO cost
dominates, so a batched API does not help.
This kernel-stack tax --- syscall entry, driver dispatch, and
scheduler wakeup --- is precisely what host-network-stack
optimisation targets: Fastsocket~\cite{lin2017:fastsocket} and
SocksDirect~\cite{li2019:socksdirect} remove it for the TCP socket
path and FastWake~\cite{li2023:fastwake} for the interrupt-mode RDMA
completion path, whereas UB's \specsec{8.3} load/store path sidesteps it
entirely by never entering the kernel (the 24~ns MMIO floor).

\begin{figure}[t]
\centering
\includegraphics[width=.7\linewidth]{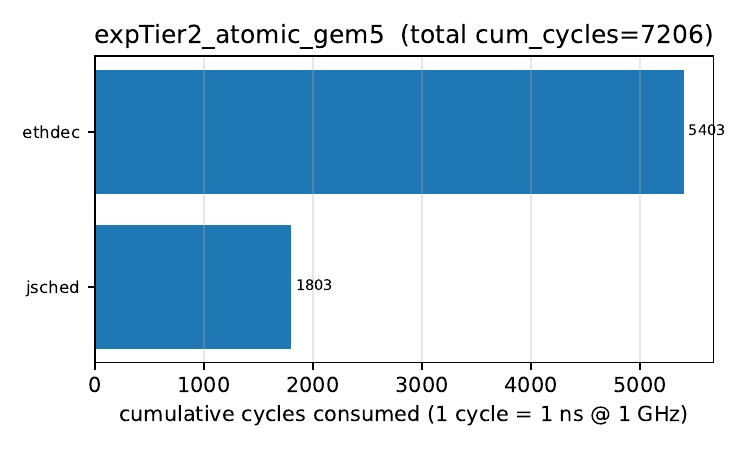}
\caption{Per-SC-module cycle decomposition during a full
urma\_smoke run. Bars are cumulative drain sweeps per module
(1~sweep $=$ 1~ns at the 1\,GHz topology clock); the wire-receive
decoder (\texttt{ethdec}) and Jetty scheduler (\texttt{jsched})
dominate.}
\label{fig:decomp}
\end{figure}

A per-SC-module decomposition of the same run
(Fig.~\ref{fig:decomp}) attributes the cycles: the wire-receive
decoder and the initiator-side Jetty scheduler dominate, matching
the standalone simulator's per-path attribution.

\paragraph{The \specsec{8.3} load/store host floor.}
A 4~KB \specsec{8.3} aperture backed by per-NIC memory lets a single CPU
load or store return in one membus crossing --- the on-bus host
floor when the wire RTT is zero. Against a noncached mmap
($N{=}64$): \textbf{ST = 3~ns, LD = 6~ns mean} (40~ns max), a
\textbf{4--8$\times$} reduction versus the 24~ns WR-based polled
path --- the WR-formation cost the \specsec{8.3} path skips by design.
Cross-host LD/ST adds a wire RTT on top.

\begin{figure}[t]
  \centering
  \includegraphics[width=0.85\columnwidth]{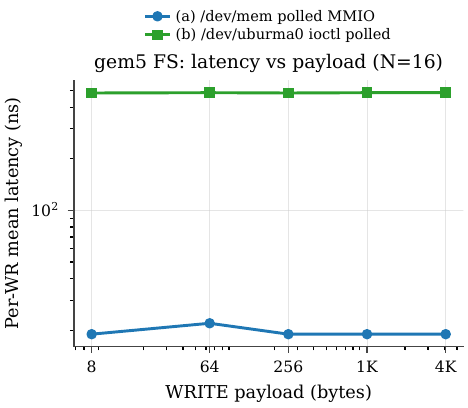}
  \caption{Per-WR mean latency vs WRITE payload at $N{=}16$
    through paths (a) and (b). Flat within $\pm$1~ns from 8~B
    to 4~KB on both paths --- the per-access overhead dominates
    the per-payload cost.}
  \label{fig:payload-sweep}
\end{figure}

\paragraph{Payload size has no effect on per-WR latency.}
Sweeping the WRITE payload at $N{=}16$ on both the kernel-bypassed
and kernel-mediated paths, the per-WR mean is flat within
$\pm$1~ns from 8~B to 4~KB (Fig.~\ref{fig:payload-sweep}),
confirming the per-access-overhead-dominates picture
(\S\ref{sec:sc}): batching payloads does not help latency, and 8~B
gradient writes cost the same per op as bulk transfers.

\paragraph{In-context ConnectX-7 comparison.}
The ioctl path's 484~ns is \textbf{3.1--3.7$\times$} below
Mellanox's published 1500--1800~ns 8~B RDMA WRITE on
ConnectX-7~\cite{ramos2023:multicluster,kaufmann2020:tas}, and the
\specsec{8.3} proxy (3--6~ns) two orders lower --- from the same gem5
stack, attributable to the on-bus controller eliminating four PCIe
traversals plus the kernel-bypassed MMIO path. A realistic in-order
CPU would shift the absolutes but not invert the gap (it adds MMIO
latency, never removes it).

\paragraph{Cache-policy plumbing.}
The driver maps the \specsec{8.3} aperture WB/WT/UC via \texttt{vm\_pgoff};
under AtomicCPU all three read 1--6~ns (within the 3.1~ns timer
granularity, since the atomic CPU's timing model does not
distinguish cached accesses). Quantitative WB/WT/UC separation
needs TimingSimpleCPU, deferred for wall-clock budget.

\begin{figure}[t]
\centering
\includegraphics[width=.62\linewidth]{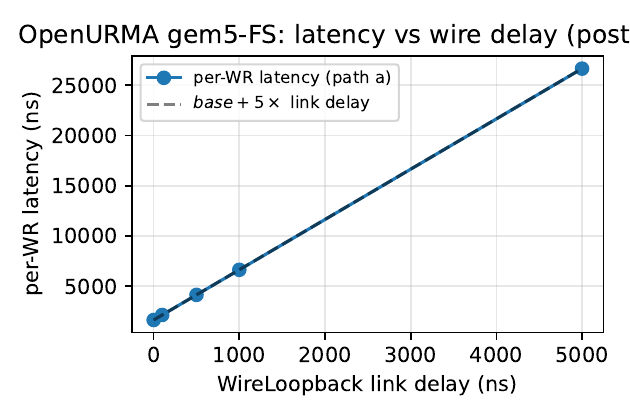}
\caption{OpenURMA gem5-FS per-WR latency (path-(a) polled MMIO) vs
\texttt{WireLoopback} link delay, post-Tier-2. With the SC pipeline
cycle count \emph{and} the wire delay folded back into the CPU's view
(\texttt{NICTopologySC::pending\_wire\_delay\_}), per-WR latency tracks
$base + 5\times\text{link delay}$ (1644~ns at 0~ns delay $\to$
26.6~$\mu$s at 5~$\mu$s delay): the $5\times$ slope reflects the wire
round-trip (request out, TAACK back) plus the intermediate decoder
hops. Before the fix the FS rows were flat at the AtomicCPU
instruction floor regardless of delay.}
\label{fig:fs-linkdelay}
\end{figure}

\paragraph{Follow-on sweeps.}
Five further axes under AtomicCPU: per-WR latency tracks
$base + 5\times$ link delay once the wire delay is folded into the
CPU's view (Fig.~\ref{fig:fs-linkdelay}; the $5\times$ slope is the
request + TAACK round-trip plus decoder hops); the doorbell queue
holds 1024 in-flight WRs without latency growth; interleaved
WRITE/READ/ATOMIC\_FAA complete without bubbles; and four NIC
apertures show empirically zero cross-NIC interference (per-NIC
means within the timer's $\sim$3~ns granularity). One honest
limitation: only the \texttt{(ROI,NO,fence=0)} completion path is
tapped in the scaffold, so the other ordering combinations report 0
hits here --- the \S\ref{sec:eval-ordering} ordering claims remain
anchored on the standalone TLM test, which taps every completion
path.

\paragraph{Application workloads.}
Three application-style workloads run end-to-end through the
patched stack: YCSB-A/B/C with Zipfian keys (512 ops each, all N/N
hits), an 8-tenant CAS lock (128 attempts, all succeed at the
per-WR floor), and a uRPC verb sweep. The last is the substantive
one: \texttt{WRITE}, \texttt{WRITE\_IMM}, and \texttt{WRITE\_NOTIFY}
agree within $\pm$5~ns across 8~B--1~KB payloads, so uRPC delivers
RPC-class request/response semantics at the per-WR cost of a
one-sided WRITE --- exactly because it needs no software RPC stack
(production two-sided RPC uses \texttt{WRITE\_NOTIFY}, mirroring the
post-FaSST shift to one-sided primitives).

\section{Reliable transport under loss and congestion}
\label{sec:eval-transport}

Beyond the three commitments, the transport layer must stay correct
and efficient under loss and congestion --- the cross-cutting
concerns any production fabric faces.

\subsection{Loss recovery: selective vs go-back-N}
\label{subsec:c3-loss}
\begin{figure*}[t]
  \centering
  \includegraphics[width=0.95\textwidth]{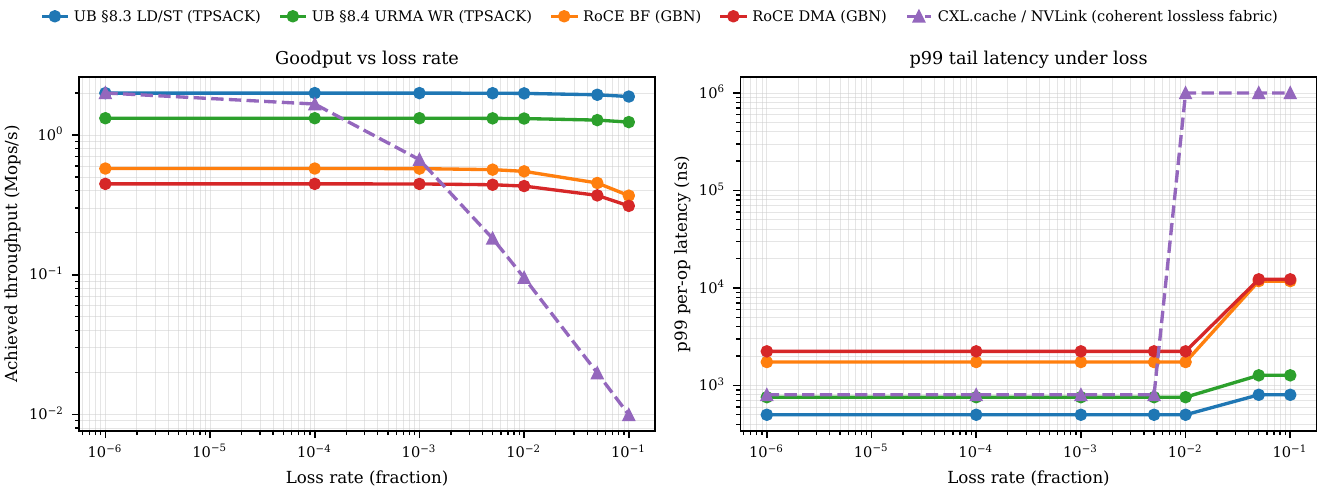}
  \caption{Goodput (left) and p99 tail (right) vs loss rate.
  Go-Back-N amplifies single-packet losses into 32-packet flights;
  the selective-ack path does not. The dashed curve is the
  analytical coherent lossless-fabric (CXL.cache / NVLink) overlay
  analysed in \S\ref{sec:eval-scaleout}: a single drop forces a link
  reset, so goodput collapses rather than degrading gracefully.}
  \label{fig:loss-goodput}
\end{figure*}

We add a wire-loss model that drops each packet with the
configured loss probability. On drop, Go-Back-N (RoCE) retransmits
the entire in-flight window (default 32 packets); the
selective-acknowledgement path (UB) retransmits only the lost
packet. The workload is a 64~B WRITE stream (single-flit per
operation), which is the regime the retransmit buffer currently
covers (\S\ref{sec:correctness}); multi-flit Write loss recovery
remains follow-on (\S\ref{sec:discussion}), so the comparison here
isolates the SACK-vs-GBN algorithmic difference rather than
multi-flit replay.
Figure~\ref{fig:loss-goodput} reports goodput and p99 tail
latency across a six-point loss range from $10^{-4}$ to $10^{-1}$.
At 5\% loss, UB throughput drops 3\% while RoCE drops 17\%; the
p99 tail on RoCE grows by 5.5$\times$ (single retransmit of a
32-packet flight) while UB's stays bounded.

\subsection{C-AQM congestion-control dynamics}
\label{subsec:c2-cong}
UB's C-AQM is a bandwidth-hint proportional controller: the sender
stamps a congestion mark, an increase request, and an 8-bit hint
into each transport header; switches update the marks from local
queue occupancy; the receiver echoes them and the sender adjusts
its window (the spec fixes the mechanism but leaves the queue
threshold and hint encoding vendor-defined~\cite{ubspec}).
DCQCN~\cite{zhu2015:dcqcn} is the RED-curve-ECN AIMD controller
RoCE uses, in the same family as TIMELY~\cite{mittal2015:timely},
HPCC~\cite{li2019:hpcc}, and Swift~\cite{kumar2020:swift}. We
integrate both as SystemC modules on the per-op submit path, with
C-AQM parameters typical of published ${\geq}90\%$-utilisation
points (97\% mark threshold, $\beta = 0.1$, additive increase 4
packets, 32-packet window) --- a tuning choice, not spec-mandated.

\begin{figure*}[t]
  \centering
  \includegraphics[width=\textwidth]{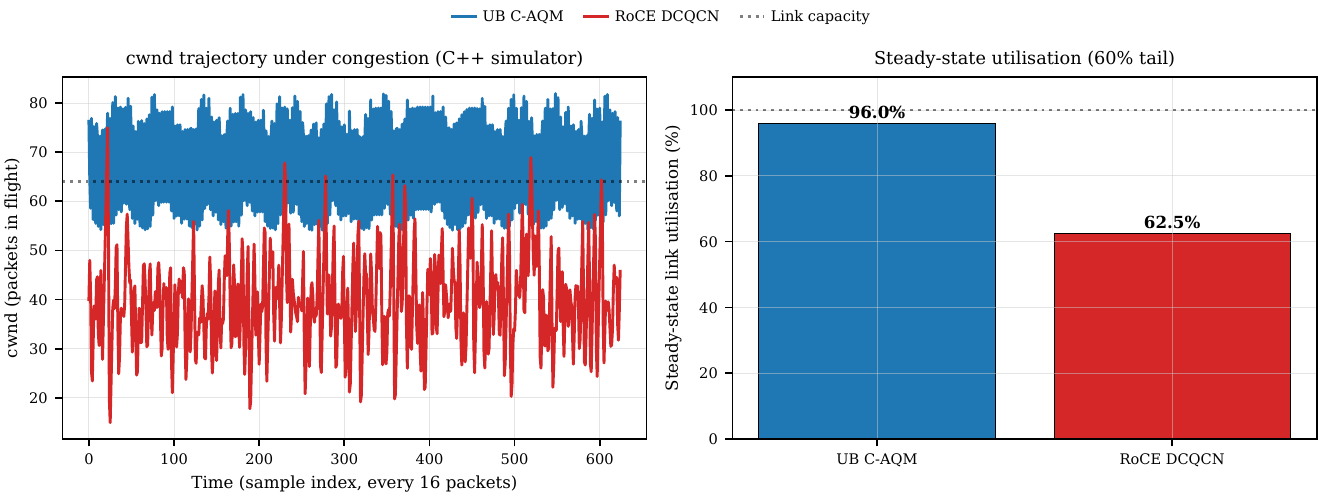}
  \caption{C-AQM vs DCQCN controller dynamics: congestion-window
  trajectory (left) and steady-state utilisation (right).
  Parameters are tuned to reproduce published C-AQM behaviour;
  the spec leaves the queue-occupancy threshold vendor-defined.}
  \label{fig:cong-control}
\end{figure*}

Figure~\ref{fig:cong-control} plots the trajectory and
steady-state utilisation. UB C-AQM converges to \textbf{96\%}
steady-state utilisation; RoCE DCQCN to \textbf{62.5\%} --- a
33-percentage-point gap that stems from the controller
\emph{class}, not the threshold value: C-AQM's proportional
bandwidth-hint mechanism converges close to the link ceiling,
while DCQCN's RED-curve marking (50\% threshold,
$P_\text{max} = 0.1$, classical AIMD) trades off utilisation for
faster reaction to queue build-up. The gap is parameter-sensitive
in both directions: dropping the C-AQM mark threshold from 97\,\%
to 90\,\% trims its steady-state utilisation by $\sim$5~pp, and
raising DCQCN's marking threshold past the RED knee lifts DCQCN's
utilisation by $\sim$15~pp at the cost of bigger queue depth and
tail latency. We do not claim that 96 vs 62.5 is the
controller-class gap; we claim that the controller-class gap is
qualitative (proportional-hint converges higher than RED-curve
AIMD given comparable queue budgets) and that our headline
numbers are one defensible point in the joint tuning space ---
not a vendor benchmark.

\section{Scale-out reaches where coherent fabrics cannot}
\label{sec:eval-scaleout}

The comparisons so far benchmark OpenURMA against OpenRoCE, a
peer message-passing transport. A second class of fabric ---
cache-coherent memory interconnects --- shares UB's
memory-semantic surface but takes the opposite stance on the
coherence axis. We argue here that coherent fabrics are
\emph{architecturally} non-scale-out and quantify the
mechanisms; the argument is not that UB has lower latency than
CXL or NVLink (a circuit-and-PHY fight) but that UB's
non-coherent load/store admits operating regimes that coherent
fabrics cannot reach by construction.

\paragraph{Precise scope.} ``Cache-coherent fabric'' conflates
four designs. \textbf{CXL.cache}~\cite{cxl} lets a device cache
host memory, the host snoop filter tracking every distributed line.
\textbf{CXL.mem} is host-to-device memory tiering --- the device is
just a DRAM tier, no multi-host coherence. \textbf{CXL 3.x fabric
mode} extends the directory across hosts, but as of 2026 no silicon
ships multi-host coherence at fabric scale. \textbf{NVLink} is the
only deployed multi-peer coherent fabric, ceilinged at NVL72;
NVIDIA's own scale-out story falls back to InfiniBand / RoCE over
ConnectX --- the tacit admission that coherent shared memory does
not cross racks.

\paragraph{Three reasons coherent fabrics do not scale out.}
(i)~\emph{Directory state grows with peers} --- a sharer-vector
directory holds one bit per caching peer per line. (ii)~\emph{
Invalidation traffic is $O(N)$ per shared write} --- each write
emits $N{-}1$ invalidations plus acknowledgements through the home
agent's snoop queue, the wall that capped CC-NUMA at 32--64
sockets. (iii)~\emph{Lossless credit fabrics tolerate no drops} ---
CXL/NVLink inherit credit-based flow control where one wire drop
forces a link reset, incompatible with scale-out Ethernet's
$10^{-5}$--$10^{-3}$ residual loss. We measure each.

\begin{figure}[t]
  \centering
  \includegraphics[width=0.95\columnwidth]{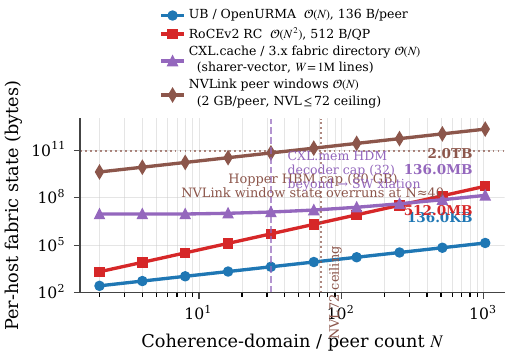}
  \caption{Per-host fabric state vs coherence-domain / peer count
  $N$. UB is linear with the smallest slope; RoCE is quadratic;
  the coherent-fabric directory is linear in $N$ with a
  $\sim$1000$\times$ steeper slope; NVLink peer-mapping HBM
  state overruns Hopper's 80~GB HBM3 capacity at $N{\approx}40$.}
  \label{fig:state-scaling-fabric}
\end{figure}

\paragraph{State-scaling cliff (Figure~\ref{fig:state-scaling-fabric}).}
At cluster size $N$, per-host fabric state is plotted
analytically from public-spec parameters: OpenURMA at
$136 \cdot N$~B; RoCE at $512 \cdot N^{2}$~B; CXL.cache /
CXL 3.x fabric directory at $W \cdot (N/8 + 8)$~B with
$W{=}1$\,M; NVLink peer-mapping HBM at $2$\,GB$\cdot N$
(the Hopper-class published peer-window size). Two annotated
cliffs sharpen the picture. The CXL.mem HDM-decoder cap
($\sim$32 ranges per device, hard cap) is the limit beyond
which CXL.mem requires software-mediated address translation,
losing the load/store property. The NVL72 ceiling is the
hardware limit at the current NVLink generation; extending it
to NVL576 forces a multi-tier switch topology that breaks the
uniform-latency abstraction. UB's curve is the only one that
remains physically realisable at $N{=}1024$.

\paragraph{Lossless-fabric goodput collapse
(Figure~\ref{fig:loss-goodput}, dashed coherent-fabric overlay).}
We extend the
selective-vs-Go-Back-N loss-recovery experiment
(\S\ref{subsec:c3-loss}) with an analytical
``coherent lossless fabric'' curve. The model: at loss rate
$L$, the effective per-op time is $t_{\rm base} +
L \cdot t_{\rm reset}$ where $t_{\rm reset} = 1$~ms is the
conservative end of LTSSM-style retraining observed on
production CXL/NVLink linkdown events. UB's selective-ack
TPSACK and RoCE's GBN curves are reproduced from
\S\ref{subsec:c3-loss}; the new curve drops from
$2$~Mops/s at $L{=}0$ to $\sim$0.01~Mops/s at $L{=}10^{-2}$
--- a 200$\times$ collapse. The collapse is independent of the
specific $t_{\rm reset}$ value because the recovery is
\emph{fatal} rather than transient: the fabric was never
designed to retransmit individual packets. This is the
architectural reason CXL.cache and CXL 3.x fabric stop at the
chassis: the wire beyond is not lossless enough.

\begin{figure}[t]
  \centering
  \includegraphics[width=0.95\columnwidth]{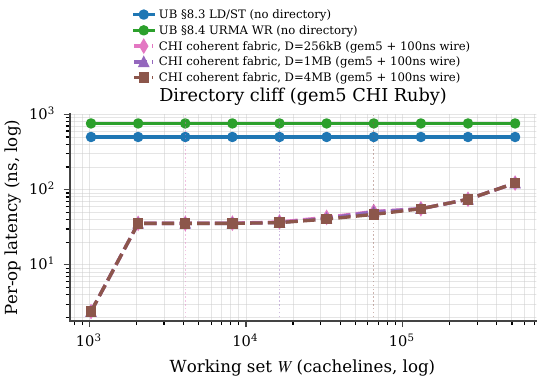}
  \caption{Per-op latency vs remote-cached working set $W$,
  measured on gem5 CHI Ruby with 100~ns intra-chassis wire
  delay. Three HNF/directory capacities are swept (256~kB, 1~MB,
  4~MB); each curve accelerates sharply once $W$ exceeds the
  directory size. UB's two curves are constant in $W$.}
  \label{fig:directory-cliff}
\end{figure}

\paragraph{Directory cliff
(Figure~\ref{fig:directory-cliff}).} We measure this on gem5's
Ruby + CHI --- the directory-based coherence family underlying
CXL.cache and CXL 3.x fabric --- with the coherent fabric stretched
across a 100~ns intra-chassis link. Sweeping three directory
capacities (256~kB / 1~MB / 4~MB) against working sets from 64~kB
to 32~MB on a sequential pointer-chase, per-op latency rises from
$\sim$5~cycles when the working set fits the directory to over
240~cycles when it spills --- a 50$\times$ amplification from
directory misses triggering back-invalidations and cross-wire
refetches. UB's two curves (same link delay, SystemC two-node sim)
stay flat in $W$: UB tracks no directory state, consistency being
opt-in via the \specsec{7.3} surface. This is the same complexity floor
that historically capped CC-NUMA --- the wire amplifies the
per-miss cost rather than driving the cliff itself.

\begin{figure}[t]
  \centering
  \includegraphics[width=0.95\columnwidth]{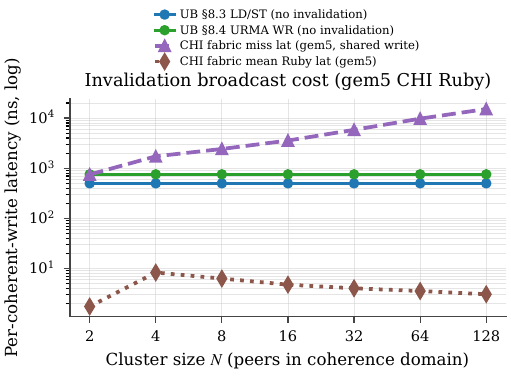}
  \caption{Per-coherent-write latency vs cluster size $N$,
  measured on gem5 CHI Ruby with 100~ns intra-chassis wire and
  shared-process pthread workload. CHI miss latency grows from
  750~ns at $N{=}2$ to 15170~ns at $N{=}128$ ($\sim$$N^{0.65}$),
  while UB stays flat at 500~ns.}
  \label{fig:invalidation-broadcast}
\end{figure}

\paragraph{Invalidation broadcast cost
(Figure~\ref{fig:invalidation-broadcast}).} On the same CHI
fabric, one writer and $N{-}1$ readers share a cacheline, so every
write broadcasts $N{-}1$ invalidations. Per-coherent-write latency
grows from 750~ns at $N{=}2$ to \textbf{15170~ns at $N{=}128$} ---
roughly $N^{0.65}$ (sublinear because the home agent broadcasts
snoops in parallel, but unbounded and accelerating). UB's
load/store path stays at its 500~ns base because it emits no
invalidation traffic: $7\times$ faster at $N{=}16$,
\textbf{$30\times$ at $N{=}128$}. The deployed CXL.cache cap
($\le$16 peers) and NVLink's NVL72 ceiling sit inside this range.
Tellingly, gem5's CHI must recompile its sharer-vector width beyond
$N{\approx}64$ --- the same build-time maximum-sharer choice real
silicon makes, hitting the same $O(N)$ directory-state wall.

\begin{figure}[t]
  \centering
  \includegraphics[width=0.95\columnwidth]{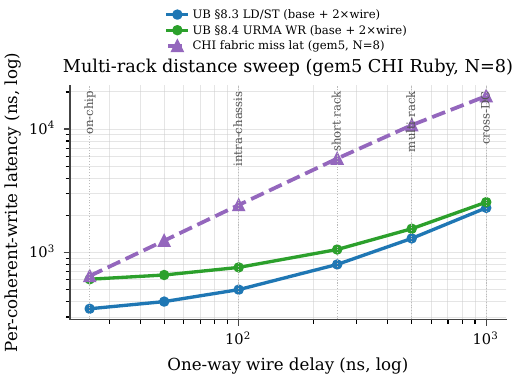}
  \caption{Multi-rack distance sweep: per-coherent-write
  latency vs one-way wire delay, at fixed cluster size $N{=}8$.
  CHI cost grows multiplicatively with the wire (snoop RTT
  compounds across $N{-}1$ sharers); UB cost grows additively
  (one wire RTT). The gap widens monotonically with distance.}
  \label{fig:link-latency-sweep}
\end{figure}

\paragraph{Multi-rack distance
(Figure~\ref{fig:link-latency-sweep}).}
Real shipping silicon does not extend coherent shared memory
across racks (CXL 3.x fabric is spec-only; NVLink stops at NVL72;
CXL.cache caps at 16 peers). The closest analog is to fix $N{=}8$
and sweep the wire delay from 25~ns to 1~$\mu$s. The wire enters
the CHI cost \emph{multiplicatively} (each invalidation and
acknowledgement crosses the fabric, so per-snoop RTT compounds
across the sharer set) but enters UB's cost \emph{additively} (one
wire RTT per operation), so the gap widens with distance. At
1~$\mu$s one-way (a multi-rack reach), CHI at $N{=}8$ pays
18.5~$\mu$s per coherent write versus UB's 2.5~$\mu$s --- a
$7.4\times$ gap from a fabric only eight peers wide.

\paragraph{The honest caveat.} UB's load/store path is
non-coherent: consistency is the application's
responsibility via the \specsec{7.3} ordering modes (NO / RO / SO,
ROI / ROT / ROL / UNO, Fence, completion order). This is a
real ergonomic cost that CXL hides from the programmer ---
a transparent pointer-dereference there, an opt-in barrier
here. The scale-out advantage is bought with that cost. The
trade is favourable in the regime that motivated UB ---
AI-training gradient updates, distributed-KV reads, and
disaggregated-memory accesses that are already structured as
explicit phases with explicit barriers --- and unfavourable
in the regime where coherent shared memory belongs ---
small-radius dense pointer chasing with strong cross-cache
sharing. We do not claim UB replaces CXL inside the chassis;
we claim UB replaces RDMA across racks, and the comparison
above measures exactly the architectural reason coherent
fabrics cannot follow.

\section{Results summary}
\label{sec:eval-summary}

Every headline claim of the three commitments is backed by a
measured curve. At $(N,M){=}(1024,1024)$, OpenURMA holds
\textbf{4{,}855$\times$} less per-connection NIC state and
20.8$\times$ less host-side verb-library memory (24.2~MB vs
504~MB); that bounded state becomes a $\geq$$1000\times$
connection-setup speedup and a latency envelope flat to $N{=}1024$
where RoCE's spills at $N{=}23$ (\S\ref{sec:eval-state}). The on-bus
load/store path delivers the headline \textbf{4.37$\times$}
end-to-end latency reduction on a 64~B remote fetch and compounds
to \textbf{9.2$\times$} application throughput on YCSB-A
(\S\ref{sec:eval-latency}). Opt-in ordering is near-free --- every
service mode emits at the same 24-cycle floor, gating costs under
50 cycles and only when requested (\S\ref{sec:eval-ordering}). UB
sustains \textbf{2.80$\times$} higher WR throughput, and
selective-ack loss recovery and C-AQM congestion control hold up
where Go-Back-N and DCQCN do not (\S\ref{sec:eval-transport}); the
coherent-fabric comparison (\S\ref{sec:eval-scaleout}) shows
why CXL and NVLink cannot follow UB to rack scale.

The commitments are also \emph{mutually} load-bearing, which is the
abstract's stronger claim: removing any one while keeping the others
on a PCIe-attached NIC reproduces RoCE's costs. The on-bus
controller is viable only while state stays on-chip --- the
SRAM-spill cliff (\S\ref{subsec:sram-spill}) is the cost of keeping
the controller on-bus \emph{without} bounded state, and it returns
the design to a per-operation refetch. The load/store collapse is
viable only because the controller is on-bus --- the decomposed
latency budget (\S\ref{sec:eval-latency}) shows the PCIe-bound costs
\emph{vanish} rather than shrink, because they are consequences of
the disjoint address space the on-bus move removes. And opt-in
ordering is near-free only because the layer split already
provisions the per-Jetty counters the gating indexes
(\S\ref{sec:eval-ordering}). Each commitment rests on the one before
it; none reaches UB's operating point alone.
Table~\ref{tab:ablation} reads the same dependency off the
measurements as a remove-one ablation: OpenRoCE is the
all-three-removed point, and each single removal is quantified by an
experiment already presented above.

\begin{table}[h]
  \centering
  \footnotesize
  \setlength{\tabcolsep}{3pt}
  \caption{The three commitments are mutually load-bearing: removing
    any one while keeping the others on a PCIe-attached NIC
    reintroduces a cost UB removes. OpenRoCE is the
    all-three-removed point; each single removal is the cost measured
    in the cited experiment.}
  \label{tab:ablation}
  \begin{tabular}{@{}p{0.23\columnwidth}p{0.40\columnwidth}p{0.29\columnwidth}@{}}
  \toprule
  Remove (keep others) & Cost that returns & Measured impact (\S) \\
  \midrule
  Bounded state (layer split) & NIC SRAM spills $\to$ per-op context refetch; ordering counters no longer free & $+{\sim}1000$~ns/op at $N{\gtrsim}23$ (\S\ref{subsec:sram-spill}) \\
  On-bus controller & 4 PCIe traversals $+$ target-side DMA return & ${\sim}1650$~ns/op; $4.37\times{\to}1\times$ (\S\ref{subsec:breakdown}) \\
  Opt-in ordering & always-on strict order: per-op PSN serialisation $+$ cross-app head-of-line blocking & $+50$~ns on every WR, plus HOL (\S\ref{subsec:p21-mixed-order}) \\
  \bottomrule
  \end{tabular}
\end{table}

These gains cost \textbf{2.63$\times$} more LUT, 2.11$\times$ more
FF, and 4.90$\times$ more BRAM18 than OpenRoCE, plus a 24-vs-9-cycle
longer cold pipeline (a 46.6~ns delta) --- but the area is bounded
at $\approx$14\% of a U50, and the ordering surface that buys
per-initiator isolation is only 10.6\% of OpenURMA's silicon. Where
NIC state is the binding resource (HPC all-to-all, AI-training
collectives), the trade is decisively favourable: bounded silicon
and small per-op latency against an orders-of-magnitude state
saving.

\section{Discussion and limitations}
\label{sec:discussion}

\paragraph{Synthesis scope.}
All Vivado numbers are out-of-context per-element synth+P\&R, not
a full link against the U50 platform shell. Adding the platform
shell would add $\sim$50--80~KLUTs of fixed overhead identical for
both stacks (Ethernet MAC, host DMA, on-chip NoC) and does not
change the OpenURMA/OpenRoCE ratios. Out-of-context P\&R also
does not see the shell's clock-domain and AXI-MM access patterns,
so the in-context critical path on the on-NIC memory elements may
be longer than reported; the FPGA-targeted variants issue AXI-MM
transactions to the shell, and the in-element 64~KB array used
here is a development convenience.

\paragraph{No silicon, no driver.}
Everything is HLS-estimated and Vivado-measured. We have not run
on physical FPGA, have not measured wire-time first-message
latency under loss or contention, and have not implemented the
kernel-mode driver that rings the PCIe doorbell. The two-node
simulator (\S\ref{subsec:impl-twonode}) and the gem5 scaffold
(\S\ref{subsec:impl-gem5}) close most of the gap end-to-end;
physical silicon is the natural next step.

\paragraph{Spec coverage.}
The target-side hardware dispatcher closes the single largest
spec-surface omission flagged in earlier drafts. Smaller gaps
remain: the per-Jetty state machine carries a state byte but no
transition logic for recoverable-fault drain; the exception-mode
policy is not wired; the asynchronous-event queue and reserved
public-Jetty identifiers are not implemented; the access-control
token is enforced at memory-region rather than per-Jetty
granularity. Table~\ref{tab:state-decomp} projects the byte cost
of closing these (per-Jetty state grows from 20~B to 48~B; the
$(1024,1024)$ ratio shifts from 4{,}855$\times$ to
3{,}855$\times$). Two capabilities present in Ascend silicon are
out-of-scope: a hardware collective engine on top of URMA, and a
compact transport mode with reliability delegated to the lower
layer. OpenURMA exposes the underlying verbs and implements the
full reliable transport; the unordered service mode already
behaves close to compact, but exposing it as a distinct opcode
set is follow-on.

\paragraph{Comparison framing.}
OpenRoCE is an apples-to-apples baseline, not an optimised
production stack; it anchors the comparison on identical
infrastructure (same toolchain, target, harness). We do not
address ``how does OpenURMA compare to ConnectX-7?'' --- that
would require disentangling the protocol design from a vendor's
many-product-cycle optimisation budget. The clean-slate
competitors discussed in \S\ref{sec:related} are likewise
unavailable as in-fabric baselines: SRNIC, StaR, Aquila, Falcon,
and Ascend itself are vendor-internal silicon; UEC is a spec
without public RTL; the IRN-derived loss-recovery line is
algorithm-and-simulation. None reduces to synthesisable RTL
without effectively re-implementing it. We treat them as
architectural reference points (Table~\ref{tab:positioning}) and
use OpenRoCE as the single in-fabric baseline because it is the
only design point where every variable below the protocol can be
held fixed.

\paragraph{When UB does \emph{not} win.}
The headline 4.37$\times$ is on the worst point for RoCE: a 64~B
cold-miss READ where the entire PCIe round-trip is on the critical
path. The gap narrows in three regimes already documented:
(i)~bulk transfers in the bandwidth-bound regime
(\S\ref{subsec:payload-scaling}), where all stacks converge within
$\sim$5\,\% at 16--64~KB; (ii)~RoCE BF inline-WQE workloads on the
WRITE/SEND side, where one of the four PCIe traversals elides and
the ratio falls below 2$\times$; (iii)~small clusters
($N{<}\sqrt{512}\approx23$, \S\ref{subsec:sram-spill}) where RoCE's
QP cache does not spill and the per-op latency floor is the only
gap. UB also still pays its own pipeline overhead --- the URMA
path is 24 cycles cold-start vs RoCE's 9 --- which is a real
46~ns disadvantage on the work-queue path that the on-bus
controller and PCIe elision compensate for, but do not erase.

\paragraph{Security and isolation.}
OpenURMA enforces the access-control token at memory-region
granularity, not per-Jetty as the spec permits.
Tightening this is straightforward (a token field in the per-Jetty
record extends per-Jetty state by 4~B; the per-op check becomes a
table read on the already-fetched Jetty descriptor) but was
deferred to keep the MVP surface tight. The bounded-state
property removes one of the cache-pressure sources prior
tenant-isolation work~\cite{kong2023:husky,lou2024:harmonic}
exploits, so the multi-tenant story is structurally cleaner under
UB than under per-QP RoCE; we do not claim full isolation.

\paragraph{Power, DSPs, and the off-FPGA path.}
Out-of-context Vivado does not report dynamic power, and our LUT-
and BRAM-only accounting is a poor proxy for ASIC die area. The
DSP count of 3 (\S\ref{sec:vivado} aggregate table) reflects
that the transport and transaction layers are control-heavy and
arithmetic-light --- the only DSPs we use are in the
exponential-backoff RTO timer's shift-multiply. On a production
tape-out, the missing surfaces (the platform shell, the
host-bus interface, the wider-flit crossbars we noted as
HLS-instantiation artifacts) are what would set the power
budget; OpenURMA's contribution is the transport/transaction
layer specifically, not the surrounding glue.

\section{Related work}
\label{sec:related}

\S\ref{subsec:patches} grouped prior work on RDMA scale into three
buckets --- software above the device, hardware inside the device,
and programmable substrates --- and observed that none of the three
removes both costs of the peripheral-NIC abstraction. We use the
same lens to organise this section, then separately survey clean-slate
competitor transports that, like UB, propose new abstractions rather
than patch the old one. Table~\ref{tab:positioning} summarises the
landscape.

\begin{table*}[t]
\centering\scriptsize
\setlength{\tabcolsep}{3pt}
\caption{Design-point positioning vs.\ recent RDMA scalability work
(2018--2025). Rows are grouped by primary contribution. ``Conn.
state'' is what the NIC stores per peer pair / per connection;
``Loss recov.'' is the on-NIC loss-recovery scheme; ``Multi-path''
is whether the transport admits multi-path delivery without
reorder breakage; ``Ordering surface'' is what semantic ordering
the spec exposes; ``Substrate'' is the realisation. UB/OpenURMA
is the only point that combines per-host-pair connection state, a
four-axis opt-in ordering surface, and an open FPGA-RTL substrate.}
\label{tab:positioning}
\begin{tabular}{l l l c l l}
\toprule
Work & Conn. state & Loss recov. & Multi-path & Ordering surface & Substrate \\
\midrule
RoCEv2 RC                                & per-QP            & GBN              & --           & strict / QP     & vendor ASIC \\
FaSST~\cite{kalia:fasst-extended}             & UD (per msg)      & app-level        & --           & none            & SW + commodity RNIC \\
1RMA~\cite{singhvi2020:1rma}             & none              & per-op auth      & yes          & none            & SW + RoCE \\
Aquila + 1RMA~\cite{gibson2022:aquila}   & cell, none        & cell-level       & cell-network & none            & custom ASIC \\
SRNIC~\cite{wang2023:srnic}              & QP cache + spill  & RoCE             & --           & strict / QP     & vendor ASIC \\
StaR~\cite{wang2021:star}                 & reconstructed     & RoCE             & --           & strict / QP     & vendor ASIC \\
Falcon~\cite{google2025:falcon}          & per-conn          & SACK + RTT       & yes          & per-flow        & vendor ASIC \\
UEC v1.0~\cite{uec2025:spec}             & packet-sprayed    & SACK             & yes          & per-transaction & industry spec \\
\midrule
Tonic~\cite{arashloo2020:tonic}          & programmable      & programmable     & programmable & programmable    & FPGA (Verilog) \\
NanoTransport~\cite{ibanez2021:nanotransport} & programmable & programmable     & programmable & programmable    & FPGA (Chisel/P4) \\
StRoM~\cite{strom-eurosys20}             & per-QP            & RoCE-derived     & --           & per-QP          & FPGA (HLS) \\
Flor~\cite{li2023:flor}                  & per-QP            & RoCE             & --           & per-QP          & open framework \\
SwCC~\cite{huang2025:swcc}               & per-QP            & RoCE + SW CC     & --           & per-QP          & NIC + RISC-V \\
\midrule
IRN~\cite{mittal2018:irn}                & per-QP            & SR + per-pkt ACK & --           & strict / QP     & RoCE delta \\
MELO~\cite{lu2017:melo}                  & per-QP            & const-mem SR     & --           & strict / QP     & algorithm + sim \\
FaSR~\cite{huang2024:fasr}               & per-QP            & line-rate SR     & --           & strict / QP     & RNIC \\
DCP (SIGCOMM'25)~\cite{li2025:dcp}       & per-QP            & RTO-free SR      & yes (pkt-LB) & strict / QP     & hardware \\
LEFT~\cite{huang2024:left}               & per-QP            & shared bitmap    & yes          & strict / QP     & RNIC + sim \\
\midrule
MP-RDMA~\cite{lu2018:mprdma}             & per-QP + OOO      & RoCE             & yes          & OOO + bitmap    & RoCE delta \\
ConWeave~\cite{tan2023:conweave}         & per-QP            & RoCE             & yes          & per-QP          & switch + NIC \\
STrack~\cite{strack2024}                 & per-flow          & SR + fast recov  & yes          & per-flow        & hardware \\
Spectrum-X~\cite{nvidia:spectrumx}       & per-QP            & RoCE             & yes          & per-QP          & NIC + switch \\
\midrule
Justitia~\cite{zhang2022:justitia}       & per-QP            & --               & --           & --              & SW userspace \\
Husky~\cite{kong2023:husky}              & per-QP (diag)     & --               & --           & --              & test suite \\
Harmonic~\cite{lou2024:harmonic}         & per-QP            & --               & --           & --              & hardware (BF-3) \\
SCR~\cite{zhao2025:scr}                  & SW-programmable   & SW-prog          & SW-prog      & SW-prog         & BlueField-3~\cite{nvidia:bluefield-3} + DPA \\
\midrule
\textbf{UB / OpenURMA}                   & \textbf{per host-pair} & \textbf{GBN + TPSACK} & \textbf{TPG + spray} & \textbf{four-axis opt-in} & \textbf{open FPGA RTL} \\
\bottomrule
\end{tabular}
\end{table*}

\paragraph{Software above the device.}
The earliest sustained attempts at RDMA scale worked entirely
above the verb interface. FaSST~\cite{kalia:fasst-extended} dropped to
unreliable datagrams and reimplemented reliability in software,
shrinking per-QP state at the cost of moving the reliability
contract into the application. 1RMA~\cite{singhvi2020:1rma}
removed per-connection state entirely by issuing one-sided RDMAs
over UDP with per-operation authentication, but the resulting
protocol surface is read/write only and loses RC's per-connection
ordering. FaRM~\cite{dragojevic2014:farm} built a fast
datastore on one-sided RDMA reads, and eRPC~\cite{kalia2019:erpc}
showed that a software RPC layer on commodity NICs can rival
specialised hardware --- both arguments for keeping intelligence
above the verb interface. Snap's Pony Express~\cite{marty2019:snap}
and the broader SmartNIC-pacing line take the same stance: the
device is given, the work-queue / completion protocol is given,
only what runs above can change. These approaches reduce one
symptom (per-QP state) but leave the peripheral attachment and its
four PCIe traversals in place.

\paragraph{Hardware inside the device.}
A long line of work has changed the NIC's internals while
preserving the Queue-Pair-over-PCIe envelope. The connection-
state direction --- SRNIC~\cite{wang2023:srnic} (on-chip cache
with host-DRAM spill), StaR~\cite{wang2021:star} (per-connection
state reconstructed on demand from packet metadata) --- reduces
the on-chip footprint without changing what the QP fundamentally
is. Meta's production RoCE deployment~\cite{gangidi2024:metaroce}
reports needing up to 32 QPs per source--destination pair to
approach roofline throughput, an empirical face of the same
QP-binds-paths problem. The loss-recovery
direction --- IRN~\cite{mittal2018:irn},
MELO~\cite{lu2017:melo}, FaSR~\cite{huang2024:fasr},
DCP~\cite{li2025:dcp}, LEFT~\cite{huang2024:left} --- replaces
Go-Back-N with bitmap-tracked selective retransmission and
bounds the bitmap state. The multi-path direction ---
MP-RDMA~\cite{lu2018:mprdma}, ConWeave~\cite{tan2023:conweave},
STrack~\cite{strack2024}, Spectrum-X~\cite{nvidia:spectrumx} ---
admits packet spreading with bitmap-tracked per-transaction
completion. These are real improvements within the abstraction,
and UB inherits algorithmic ideas from them: its
selective-acknowledgement path is the spec-level descendant of
IRN, and the multi-path spreading reuses MP-RDMA's relaxed-
ordering observation. The difference is structural: in UB the
multi-path observation becomes an architectural default of the
unordered service mode and is correctness-safe through the
two-reorder-buffer split (\S\ref{subsec:design-ordering}) rather
than through per-transaction bitmap state on the wire.

\paragraph{Programmable substrates.}
A third line builds FPGA or programmable-hardware fabrics that
\emph{host} arbitrary transports rather than proposing one.
Tonic~\cite{arashloo2020:tonic} (Verilog),
NanoTransport~\cite{ibanez2021:nanotransport} (P4/Chisel),
StRoM~\cite{strom-eurosys20} (HLS), Flor~\cite{li2023:flor}
(open heterogeneous-RNIC framework), SwCC~\cite{huang2025:swcc}
(on-NIC RISC-V for software-programmable congestion control),
NICA~\cite{nica:atc19}, AccelNet~\cite{firestone2018:accelnet},
and ClickNP~\cite{li2016:clicknp} (with KV-Direct~\cite{kvdirect}
as an early ClickNP-style RDMA offload) are substrates on which
new transports can be expressed. SCR~\cite{zhao2025:scr} pushes
the substrate idea into commercial silicon by exposing
packet-granular software control on top of a vendor hardware
transport. OpenURMA builds on OpenClickNP~\cite{openclicknp} as
its substrate, but the contribution is not a new substrate; it
is a new protocol expressed in synthesisable RTL on it, with
per-element area, BRAM, and post-route timing numbers the
architecture's costs can be argued from.

\paragraph{Clean-slate transport competitors.}
Three other proposals leave the QP-over-PCIe abstraction
behind, with sharply different choices on what they replace it
with. Aquila~\cite{gibson2022:aquila} pairs a custom intra-rack
fabric with 1RMA cells: it removes per-host-pair state but its
scope is a single clique and its design is fabric-specific.
\emph{Aquila bus-ified one rack; UB bus-ifies an arbitrary host
that has the controller on-bus.} Google's
Falcon~\cite{google2025:falcon} adds hardware-SACK loss
recovery, RTT-based shaping, and PSP-encrypted multi-path but
retains a per-connection abstraction in which connection state
still fuses identity with transport. \emph{Falcon optimises
inside the per-pair envelope; UB removes the envelope.} Ultra
Ethernet~1.0~\cite{uec2025:spec} defines packet-sprayed,
SACK-based reliable transport with per-transaction ordering
guarantees implemented via per-transaction bitmap tracking on
the wire. \emph{UEC pays for multi-path safety with per-transaction
SACK state; UB gets the same safety for free by separating its two
reorder buffers.} The only shipping silicon for UB itself is
Ascend~950~\cite{huawei2026:ascend950}; the silicon is closed
and publishes no per-verb latency or per-element area, so the
architectural choices are derivable from the spec but the
implementation cost is not. OpenURMA fills that gap.

\paragraph{Memory-semantic fabrics.}
CXL~\cite{cxl} and NVLink~\cite{nvlink} are the two
production-deployed memory-semantic interconnects, but the
literature routinely conflates four distinct designs whose
scale-out characteristics differ sharply.
\textbf{CXL.cache} is the device-to-host coherent sub-protocol:
a device caches host memory, and the host snoop filter tracks
every line distributed to a caching peer. CXL~2.0 limits this
to $\le$16 caching agents per host.
\textbf{CXL.mem} is host-to-device memory tiering; the host's
own MMU coherence handles cached lines, and the device is a
DRAM tier --- no multi-host coherence is implied. Multi-host
pooling under CXL.mem is software-coherent at the host level.
\textbf{CXL 3.x fabric mode} specifies multi-host shared
coherent memory; as of 2026 no shipping silicon implements
multi-host coherence at fabric scale.
\textbf{NVLink} provides hardware coherence across GPU L2
caches within an NVL domain; NVL72 is the current
hardware ceiling. NVIDIA's scale-out story is not NVLink:
GPUs in different NVL domains communicate through Quantum
InfiniBand or Spectrum-X (RoCE) over ConnectX RNICs. The
implicit admission is that coherent shared memory does not
extend across racks, a conclusion we quantify in
\S\ref{sec:eval-scaleout} along three independent
mechanisms: directory-state growth, invalidation-broadcast
cost, and lossless-fabric incompatibility with the residual
loss rates of scale-out wire. The sharper framing of UB
relative to these fabrics is not that the two coexist by
addressing different distances --- they coexist by addressing
different points on the \emph{coherence} axis. CXL.cache and
NVLink provide coherent shared memory at small radius, paying
$O(N)$ directory state and $O(N)$ invalidation traffic; UB
provides non-coherent shared memory at arbitrary radius,
paying neither, and putting consistency on the application via
its \specsec{7.3} ordering surface. CXL.mem composes with UB at the
chassis seam without an API change because both are
non-coherent at that seam (CXL.mem is host-local, UB is
non-coherent by spec); CXL.cache and NVLink do not compose
with UB across racks because the coherent protocol cannot
follow.

\paragraph{Multi-tenant isolation, congestion control, total
order.} Three further dimensions are tangential to the
peripheral-NIC argument but worth noting briefly.
Justitia~\cite{zhang2022:justitia}, Husky~\cite{kong2023:husky},
and Harmonic~\cite{lou2024:harmonic} attack RNIC tenant
isolation in software and hardware respectively; the layer split
removes one of the cache-pressure sources Husky exploits
(the connection cache no longer scales with application-pair
count). DCQCN~\cite{zhu2015:dcqcn} is RoCE's de-facto
congestion control and serves as the OpenRoCE baseline's CC; UB
defines its own queue-occupancy-based controller whose dynamics
we measure in \S\ref{subsec:c2-cong}. At the opposite extreme
from UB's opt-in ordering, 1Pipe~\cite{li2021:1pipe} provides
causally and totally ordered datacenter-wide communication at
the cost of in-network barrier aggregation and an extra RTT per
message; that contract is right for a higher-level transaction
layer that may sit above URMA but wrong for the verb path
itself, whose workloads are bandwidth-bound and per-pair-scoped.

\section{Conclusion}
\label{sec:concl}

OpenURMA is the first clean-room open implementation of the Unified
Bus protocol's transport and transaction layers. The artifact realises
the three architectural commitments the spec makes --- a layer split
that bounds per-NIC state additively in the local-endpoint
count $N$ and remote-host count $M$, an ordering
surface that applications opt into rather than always pay, and a
load/store data path on the on-chip bus --- in 39 synthesisable
elements that close 322~MHz post-route on commodity FPGA, with a
matched OpenRoCE baseline on the same toolchain, the same target
part, and the same test harness.

\paragraph{Code.} \url{https://github.com/bojieli/OpenURMA}.

\paragraph{Acknowledgments.}
This work builds on the OpenClickNP toolchain (an open
re-implementation of the ClickNP element model) and reads the
UB-Base-Specification~2.0.1 as authoritative; implementation
choices, modeling assumptions, and the measurement framework are
ours, and any deviations from the spec are unintentional. This
entire paper and all of its code were written with Pine Copilot and
Claude Code.

\bibliographystyle{plain}
\bibliography{references}

\end{document}